\documentclass[10pt,twocolumn,letterpaper]{article}

\usepackage{iccv}
\usepackage{times}
\usepackage{epsfig}
\usepackage{graphicx}
\usepackage{amsmath}
\usepackage{amssymb}

\usepackage{multirow}
\usepackage{booktabs}
\usepackage{multirow}
\usepackage{caption}
\usepackage{subfigure}
\usepackage{diagbox}
\usepackage[pagebackref=true,breaklinks=true,letterpaper=true,colorlinks,bookmarks=false]{hyperref}

\iccvfinalcopy % *** Uncomment this line for the final submission

\def\iccvPaperID{4480} % *** Enter the ICCV Paper ID here

\ificcvfinal\fi

\begin{document}

%%%%%%%%% TITLE
% \title{Language-Guided Style Transfer Using the Latent Space of DALL-E}
\title{$\mathtt{Styler}$DALLE: Language-Guided Style Transfer Using \\ a Vector-Quantized Tokenizer of a Large-Scale Generative Model}

\author{Zipeng Xu\textsuperscript{1} \quad
Enver Sangineto\textsuperscript{2} \quad
Nicu Sebe\textsuperscript{1} \\
\textsuperscript{1}University of Trento, Italy \quad  \textsuperscript{2}University of Modena and Reggio Emilia, Italy \\
{\tt\small \{zipeng.xu, niculae.sebe\}@unitn.it \quad  enver.sangineto@unimore.it}
}

\twocolumn[{
\maketitle
\begin{center}
    \captionsetup{type=figure}
    % \vspace{-0.3cm}
    \includegraphics[width=.975\linewidth]{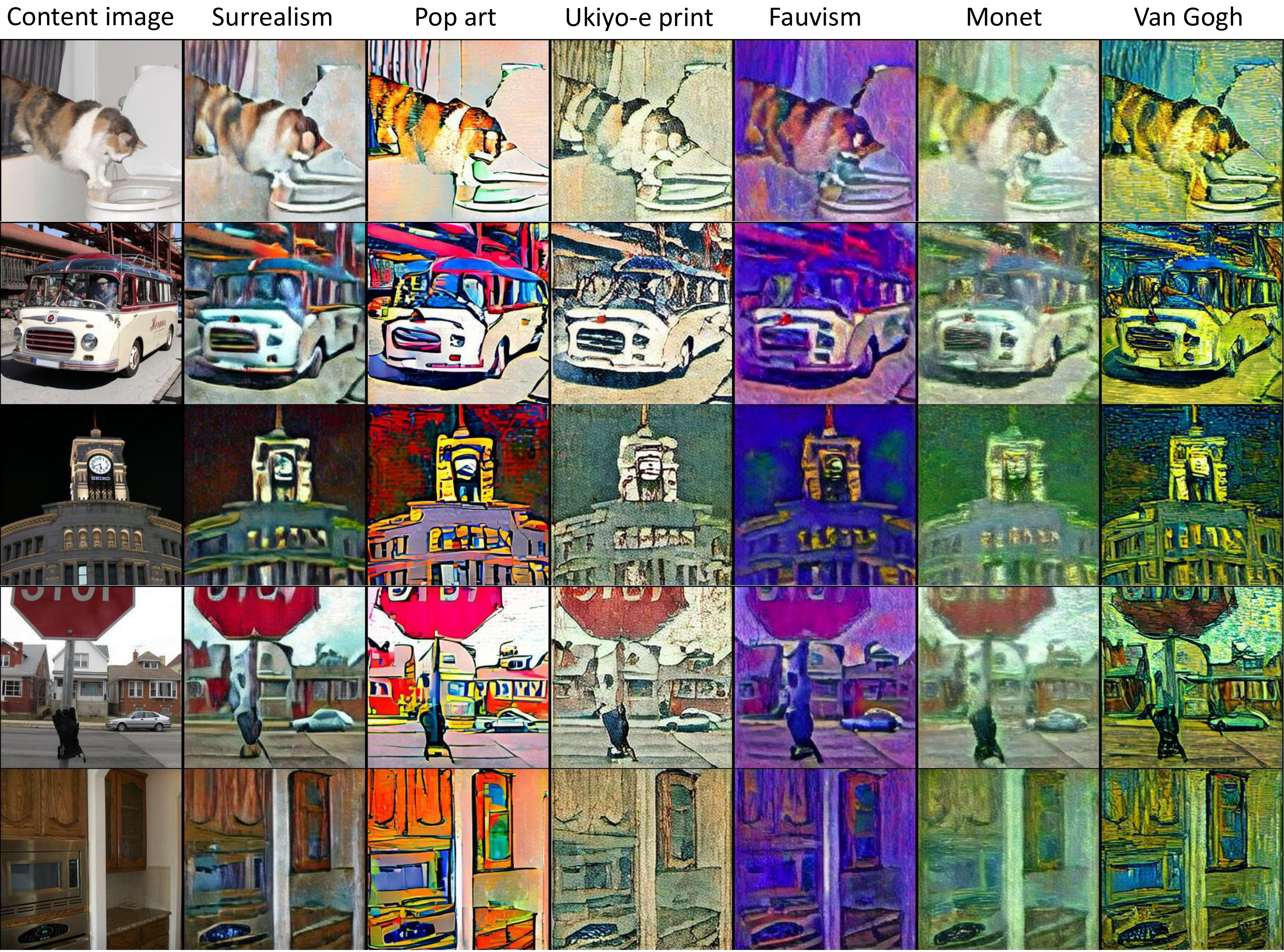}
    \captionof{figure}{With a large-scale pretrained vector-quantized tokenizer (e.g., the dVAE of DALL-E) and CLIP, $\mathtt{Styler}$DALLE can transfer various types of styles  (indicated on the top), from abstract art styles to specific artist styles and more.
    }
    \label{fig:1}
    % \vspace{-0.2cm}
\end{center}
}]

% \maketitle
\ificcvfinal\thispagestyle{empty}\fi

%%%%%%%%% ABSTRACT
\begin{abstract}
% \vspace{-0.3cm}
Despite the progress made in the style transfer task, most previous work focus on transferring only relatively simple features like color or texture, while missing more abstract concepts such as overall art expression or painter-specific traits. However, these abstract semantics can be captured by models like DALL-E or CLIP, which have been trained using huge datasets of images and textual documents. In this paper, we propose $\mathtt{Styler}$DALLE, a style transfer method that exploits both of these models and uses natural language to describe abstract art styles. Specifically, we formulate the language-guided style transfer task as a non-autoregressive token sequence translation, i.e., from input content image to output stylized image, in the discrete latent space of a large-scale pretrained vector-quantized tokenizer, e.g., the discrete variational auto-encoder (dVAE) of DALL-E. To incorporate style information, we propose a Reinforcement Learning strategy with CLIP-based language supervision that ensures stylization and content preservation simultaneously. Experimental results demonstrate the superiority of our method, which can effectively transfer art styles using language instructions at different granularities. Code is available at \href{https://github.com/zipengxuc/StylerDALLE}{https://github.com/zipengxuc/StylerDALLE}.
\end{abstract}

%%%%%%%%% BODY TEXT
\section{Introduction}
\label{sec.Introduction}
In the last few years, a lot of work has focused on the style transfer task using a {\em reference image} as a representative of the target style, where the goal is to transfer the style of the reference to a content image~\cite{gatys2016image, huang2017arbitrary, park2019arbitrary,Chen_2021_CVPR, Lin_2021_CVPR, Deng_2022_CVPR, deng2022stytr2, wang2022aesust}.
Recent improvements in this field include: reducing the  artifacts~\cite{chen2021artistic, DBLP:conf/cvpr/LiuSCBZSL0N21, cheng2021style, wang2022aesust}, modeling the style-content relationship~\cite{park2019arbitrary,Yao_2019_CVPR, liu2021adaattn}, increasing the generation diversity~\cite{li2017diversified, ulyanov2017improved, zhang2019multimodal} and many others.
However, art styles are usually abstract concepts, e.g., pop art, fauvism, and the style of Van Gogh. 
To transfer these abstract art styles to a content image, the low-level features (e.g., textures and colors), which are commonly extracted  from a single reference image, are not enough.
A possible solution is to collect a set of reference images that can be used, e.g., to train a Generative Adversarial Network (GAN) for artist-specific style transfer~\cite{sanakoyeu2018styleaware, zhu2017unpaired, kotovenko2019content}.
The disadvantage of this {\em set-based representation} is the effort required to collect sufficiently large style-specific data for training.

Recently, large-scale image generative models~\cite{DALL-E, ramesh2022hierarchical, saharia2022photorealistic, yu2022scaling} have shown their power to generate high-quality images of various types, e.g., realistic photo, cartoon, ukiyo-e print, or painting of a specific artist. 
Moreover, CLIP~\cite{radford2021learning}, which is trained with 400 million text-image pairs, learns good joint knowledge between language and vision.
Can we leverage the power of these large-scale models for style transfer?
In this paper, we propose $\mathtt{Styler}$DALLE, a language-guided style transfer method that uses both the vector-quantized tokenizer of large-scale image generative methods and CLIP.
There are three advantages of our method. Firstly, as compared to images, language is a more natural and efficient way to describe abstract art styles and enables more flexibility. Language can directly indicate art styles at different levels, e.g., \textit{``Van Gogh style oil painting"} and \textit{``Van Gogh starry night style oil painting"}. Secondly, using CLIP and the language description of style saves the effort of collecting style images, as CLIP already learns style-related knowledge and thus can be used to provide supervision. Thirdly, using a large-scale pretrained vector-quantized tokenizer potentially enables style transfer results to be closer to real artworks, as the model is trained from an enormous number of real-world images.
In concrete, we propose a  Non-Autoregressive Transformer (NAT) model \cite{NAT}, which translates tokens of a content image to tokens of a stylized image in the discrete latent space of a pretrained visual tokenizer, and a two-stage training paradigm.

First of all, since both the input content image and output stylized image can be represented by a sequence of tokens via the large-scale pretrained vector-quantized tokenizer, we formulate the language-guided style transfer task as a token sequence translation.
Specifically, we design a NAT model that learns to translate a token-based representation of a low-resolution image into a full-resolution representation, where the final token sequence contains appearance details specialized for the target style.
The most important advantage of using NAT in image generation, with respect to the more common autoregressive transformer (AT) based generation, is that NAT is much faster than AT at inference time, as it allows a  {\em parallel} token generation while AT generates in a token-by-token way.
% Training paradigm

Secondly, we propose to use CLIP-based language supervision to ensure stylization and content preservation simultaneously, saving the effort to collect data and design dedicated losses.
% a two-stage training paradigm, which includes a textual-prompt-based Reinforcement Learning strategy to incorporate style-specific information in the translation network using the CLIP space as the only guidance, saving the effort to collect data and design dedicated losses. [language advantage CLIP]
A similar solution that uses CLIP to do style transfer is CLIPStyler \cite{kwon2022clipstyler}. Since merely maximizing the CLIP similarity with respect to a textual style description is not enough for a style transfer task, which should also {\em preserve} the content of the source image, CLIPstyler introduces a hybrid of losses, including a content loss measured in an external pre-trained VGG network.
Conversely, we propose an alternative direction, which avoids the need to tune the coefficients of different loss functions so as to keep the balance between style and content.
Specifically, we introduce a two-stage training paradigm: 1) a self-supervised pretraining stage, where the model learns to add semantically coherent image details from a low-resolution image to a high-resolution image; and 2) a style-specific fine-tuning stage, where the model learns to incorporate style into the high-resolution image.
Since the fine-tuning phase is built on top of the first stage, the translator is able to keep the semantic consistency with respect to the input image as learned during pretraining. 
Moreover, we create a textual prompt by concatenating both the style and the textual description of the image content (i.e., its caption). This way, the prompt simultaneously models both the target appearance (i.e., the style description) and the image content which should be preserved by the translation process.
Finally, since our translator's output is discrete and there is no ground-truth tokens for a stylized image, we
% cannot directly backpropagate the CLIP similarity values through our network. To solve this problem, 
introduce a Reinforcement Learning (RL) approach to fine-tune the translator using a reward based on the CLIP similarity between the stylized image and textual prompt, enabling the model to explore the answers in the latent space of a pretrained vector-quantized tokenizer.
% We use the COCO~\cite{lin2014microsoft} to train the model, using images in the first stage and image-text pairs in the second stage. 

We call our network $\mathtt{Styler}$DALLE and we show that it can generate  stylized images driven by different types of language guidance.
Compared with previous language-guided and reference image-based  transfer methods, our generated images are less inclined to produce artifacts or semantic errors. Moreover, they can capture  abstract concepts related to the target style (e.g., the typical brushstrokes of the artist) besides low-level features like texture and colors. 
We illustrate the effectiveness of our method through qualitative results, quantitative results, and a user study.
% Finally, $\mathtt{Styler}$DALLE is largely domain-independent, and it can be used with basically any type of content image (e.g., animals, indoor/outdoor images, etc.).

To conclude, our main contributions are:
\begin{itemize}
    \item We propose $\mathtt{Styler}$DALLE, a language-guided style transfer method that manipulates the discrete latent space of a pretrained vector-quantized tokenizer using a token sequence translation approach.
    \item We propose a non-autoregressive translation network that translates a low-resolution content image into a full-resolution image with style-specific details.
    \item We propose a two-stage training procedure, including an RL strategy to ensure stylization and content preservation using CLIP-based language supervision.
    \item Experimental results show that $\mathtt{Styler}$DALLE can effectively transfer abstract style concepts, going beyond simple texture and color features  while simultaneously preserving the semantic content of the translated scene.
\end{itemize}

\section{Related Work}
\paragraph{Reference Image-Based Style Transfer.}
% Style  and texture transfer tasks were studied since the beginning of the 21st century~\cite{efros2001image, hertzmann2001image}.
Gatys et al.~\cite{gatys2016image} propose a neural style transfer method in which a pre-trained CNN is used to extract content and style information from images, and to transfer the latter from an image to another.
Following this pioneering work, a lot of interests have been attracted, with different methods focusing on different aspects of the topic, such as, e.g.,  diversified style transfer~\cite{ulyanov2017improved, wang2020diversified}, or attention mechanisms to  fuse  style and content~\cite{yao2019attention, park2019arbitrary,liu2021adaattn}.
A specific  line of work focuses on artistic style transfer. For instance, Chen et al.~\cite{Chen_2021_CVPR} propose to use internal-external learning and contrastive learning with GANs to bridge the gap between human artworks and AI-created artworks.  Wang et al.~\cite{wang2022aesust} introduce an aesthetic discriminator trained with a large corpus of human-created artworks.
Other works train GANs using  an artist-specific collection of images~\cite{sanakoyeu2018styleaware,kotovenko2019content, chen2021dualast}.
In contrast, we use the generic visual-language semantics embedded in the large-scale pretrained vector-quantized tokenizer and CLIP to avoid collecting style or artist-specific datasets.

% \vspace{0.2cm}
\paragraph{Language-Guided Style Transfer.}
Very recently a few works have proposed  transferring methods conditioned on a textual description of the style. For instance, 
Fu et al.~\cite{fu2022ldast} use  contrastive learning to train a GAN for artistic style transfer,
but they adopt descriptive language instructions rather than more abstract  style concepts.
Gal et al.~\cite{gal2022stylegan} use the CLIP space for a domain adaptation  of a pre-trained StyleGAN ~\cite{karras2020analyzing}.
The method closest to our approach is CLIPStyler~\cite{kwon2022clipstyler}, where 
a patch-wise CLIP loss is used %jointly with other loss functions 
 to train a U-Net \cite{DBLP:conf/miccai/RonnebergerFB15}. 
 %\zipeng{this sentence mixes two work... Previously I wrote:  \cite{kwon2022clipstyler} proposes to use patch-wise CLIP loss to train a %Convolutional Neural Network and  }.     }
However, to condition the style change while preserving the image content, CLIPStyler uses hybrid losses and a rejection threshold, introducing many hyperparameters. In our method, we only use CLIP-based language supervision to ensure the style and content, saving the efforts of designing losses and tuning hyperparameters.
% does not have critical hyperparameters and can simultaneously change the style and preserve the content by jointly using the DALL-E and the CLIP space.

% \vspace{0.2cm}
\paragraph{Large-scale  Text-to-Image Generation Models.}
Recently,  text-to-image models trained with large or very large scale datasets \cite{DALL-E, ramesh2022hierarchical, saharia2022photorealistic, yu2022scaling, chang2023muse} have attracted tremendous attention because of their excellent performance in generating high-quality images starting from a textual query.
Inspired by their strong ability to synthesize various types of images, we study how to use large-scale text-to-image models for style transfer.
Specifically, we focus on transformer-based methods, which are potentially less time-consuming as compared to diffusion model-based methods~\cite{mokady2022null}.
In addition, we propose to use a non-autoregressive transformer that can generate tokens in parallel.

\section{Background}
\label{sec.Background}
\paragraph{Vector-Quantzied Image Tokenizer.} 
Transformer-based text-to-image generative models~\cite{DALL-E, ding2021cogview, ding2022cogview2, yu2022scaling, chang2023muse} rely on a vector-quantized image tokenizer to produce
a discrete representation of the images, e.g., DALL-E~\cite{DALL-E} has a dVAE~\cite{DBLP:conf/nips/RazaviOV19} and PARTI~\cite{yu2022scaling} has a ViT-VQGAN. 
Despite the differences in implementing the tokenizers, the effects are the same.
In more detail, an image $I$ is transformed into a $k \times k$ grid of tokens $X = \{ x_{i,j} \}_{i,j=1,...k}$, %($1 \leq i,j \leq k$) 
using an encoder $E(\cdot)$. Each token $x_{i,j} \in X = E(I)$ is an index of a codebook of embeddings
($C = \{ \pmb{e}_1, ... \pmb{e}_M \}$, $1 \leq x_{i,j} \leq M$), built during the training, and corresponds to a patch in $I$.
A decoder $G(\cdot)$ takes as input a grid of embeddings and reconstructs the original image: $\hat{I} = G(\{ \pmb{e}_{x_{i,j}} \}_{i,j=1,...k})$.
Training in the transformer-based text-to-image generative models consists of two stages. The first stage is dedicated to training the image tokenizer, while in the second phase, a transformer is used to learn a  prior distribution   over the text and the image tokens. In $\mathtt{Styler}$DALLE (Sec.~\ref{sec.Method}) we only use the pretrained image tokenizer.

\begin{figure*}[t]
    \begin{center}
    \includegraphics[width=.975\linewidth]{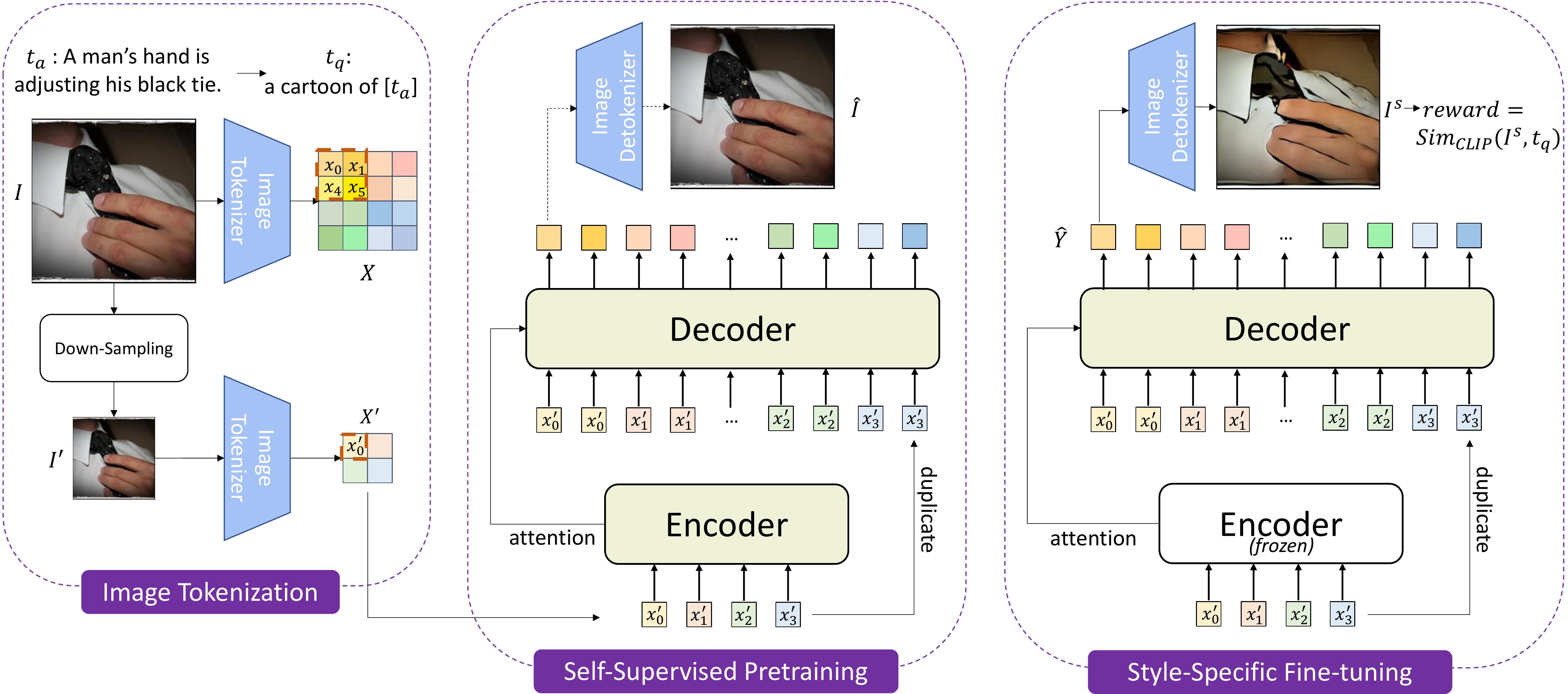}
    \end{center}
    % \vspace{-0.2cm}
    \caption{A schematic illustration of $\mathtt{Styler}$DALLE. We propose a NAT to translate tokens of a half-resolution {\em content} image to tokens of a full-resolution {\em stylized} image. $\mathtt{Styler}$DALLE consists of three parts: 1) image tokenization preprocessing, to obtain the discrete visual tokens from a pretrained image tokenizer; 2) self-supervised pretaining, to train the NAT to predict full-resolution content images from half-resolution content images; and 3) style-specific fine-tuning, to let the model adding style details in the up-scaling prediction via Reinforcement Learning with language-prompted CLIP guidance. Specifically, in our NAT, the decoder input is duplicated from the tokens of the half-resolution image, following the natural positional correspondence (as highlighted by orange rectangles in the left part) with the full-resolution image.}
\label{fig.method}
% \vspace{-0.2cm}
\end{figure*}

% \vspace{0.2cm}
\paragraph{Non-Autoregressive Transformer (NAT).}
Gu et al.~\cite{NAT} propose a NAT for natural language translation, which consists of an encoder and a decoder. The encoder  takes as input a source sentence $X = \{ x_1, ..., x_{N'} \}$ of $N'$ tokens and outputs a distribution
over possible output sentences $Y = \{ y_1, ..., y_N \}$, where $Y$ is the translation of $X$ in the target language and, usually, $N \neq N'$. 
The main novelty of NAT with respect to AT is that, during
training, NAT uses a {\em conditional independent} factorization
for the target sentence and the following log-likelihood:
% \vspace{-0.2cm}
\begin{equation}
\label{eq:NAT-loss}
% \vspace{-0.1cm}
\mathcal{L}(\theta) =\sum_{n=1}^{N} \log~p(y_n|x_{1:N'}; \theta),
% \vspace{-0.1cm}
\end{equation}
% \vspace{-0.1cm}
which differs from the common AT factorization in which the prediction of the $n$-th token ($y_n$) depends on the previously predicted tokens ($p(y_n|y_{0:n-1}, x_{1:N'}; \theta)$). This conditional independence assumption makes it possible a {\em parallel} generation of $Y$ at inference time, largely accelerating the translation time with respect to AT models. Importantly, to make parallel generation possible, the encoder input ($X$) is provided as input to the decoder as well, and  individual tokens ($x_n \in X$) can be copied zero  or more times, with
the number of times each input is copied depending on a specific predicted “fertility” value. As we will see in Sec.~\ref{sec.upscaling}, we do not need to predict fertilities because, in our case, the cardinality of the input copies is fixed and determined by the up-scaling task we use for the image translation.

\section{Method}
\label{sec.Method}
The language-guided
style transfer task  
can be described as follows. Given an image $I$, we want to generate a new image $I^s$ which preserves the semantic content of $I$ but changes its appearance  according to a textual style description $t_s$ (e.g., \textit{``cartoon"}).
In $\mathtt{Styler}$DALLE, we formulate this task as a {\em visual-token based  translation}, from tokens of a content image to tokens of a stylized image.
Specifically, given a content image $I$ we first downsample $I$ to get a half-resolution  image $I'$. Then, $I'$ is fed to the tokenization encoder (Sec.~\ref{sec.Background}) which extracts a discrete grid of $k \times k$ source tokens $X' = E(I')$. $X'$ can now be ``translated" into a target (discrete) representation $\hat{Y}$, where $\hat{Y}$ is a grid at the original resolution ($2k \times 2k$) and $\hat{Y} = f(X')$, being $f(\cdot)$ the translation function. Finally, $\hat{Y}$ is fed into the tokenization decoder obtaining the stylized image $I^s = G(\hat{Y})$.
In the following subsections, we describe the architecture of $f(\cdot)$ and the way in which it is trained.
% To solve the problem, we propose a Non-Autoregressive Transformer (NAT) model (Sec.~\ref{sec.upscaling}) and a two-stage training paradigm (Sec.~\ref{sec.upscaling} and~\ref{sec.fine-tuning}).

\subsection{Architecture and self-supervised pre-training}
\label{sec.upscaling}

For our translation network $f(\cdot)$
we use a NAT architecture \cite{NAT} (Sec.~\ref{sec.Background}) which we train from scratch using a self-supervised learning pretext task consisting in predicting the image at full resolution. Specifically, given the downsampled image $I'$ and its corresponding grid of tokens $X' = E(I')$, we use the indexes in $X'$ to extract the corresponding embeddings from $C$ (Sec.~\ref{sec.Background}). For each $x_{i,j} \in X'$, let $\pmb{e}_{x_{i,j}}$ be the corresponding embedding in $C$ and let $N' = k^2$.
The so obtained set of embeddings is flattened into a sequence, and, for each element $\pmb{e}_n$ of the sequence ($1 \leq n \leq N'$), we add an absolute positional embedding \cite{attention-is-all-you-need} $\pmb{p}_n$, where $\pmb{p}_n$ has the same dimension as $\pmb{e}_n$:
$\pmb{v}_n = \pmb{e}_n + \pmb{p}_n$. 
The final sequence $V' = \{ \pmb{v}_1, ..., \pmb{v}_n, ..., \pmb{v}_{N'} \}$ is input to the encoder of $f(\cdot)$.
Note that an alternative solution is to directly fed $f(\cdot)$ with (a flattened version of) $X'$
and let $f(\cdot)$ learn its own initial token embedding.
However, using the embeddings in $C$ has the advantage of exploiting the image representation of the pre-trained image tokenizer. 
Moreover, from the original image $I$ we extract the ground truth $X = E(I)$, which is flattened in a sequence of $N = 4k^2$ tokens.

Finally, following \cite{NAT}, we build a second sequence of input embeddings $V$, with cardinality $N$,
which is fed to the decoder of $f(\cdot)$.
As mentioned in Sec.~\ref{sec.Background}, differently from NAT, we do not predict fertility values. Instead, we get the input of the decoder by simply replicating each element $x_{i,j} \in X'$ according to the positional correspondences between the low-resolution image and high-resolution image (as in Fig.~\ref{fig.method}).  
The rationale behind this choice is that $f(\cdot)$ is trained to predict the full-resolution image, and each encoder input  ($\pmb{e}_{x_{i,j}}$) corresponds to a patch in the subsampled image $I'$ and to 
4 patches in the  full-resolution image $I$.
Thus, initializing the decoder with 4 replicas of each source-image patch initial embedding provides a coarse-grained signal for the upsampling task.
Similar to before, the embeddings extracted from $C$ are then flattened and added with a new positional encoding. 
% (computed over the new sequence of $N$ elements).
% Note that $X \neq Y$, and we cannot use $Y$ to build $V$ because this would  make the upsampling task of the decoder a trivial identity mapping operation.

Both the encoder and the decoder have self-attention layers and no causal masking is used. However, following \cite{NAT}, in the decoder, we mask out each query position ($n$) only from attending to itself.
Using $V'$ and $V$,  $f(\cdot)$ generates $N$ {\em parallel} posterior distributions over the visual vocabulary ($\{1, ..., M \}$):
$P = f_{\theta}(V', V)$, where $P$ is a $N \times M$ matrix, $P_n \in [0,1]^M$
and $P_n[y] = p_{\theta} (Y_n = y | V', V)$.
Using $Y = \{y_1, ..., y_n, ..., y_N \}$,
%(where each $y_n \in Y$ is a token of the dVAE vocabulary: $y_n \in  \{1, ..., M \}$),  
$f(\cdot)$ is trained to maximize:

\begin{equation}
\label{eq:our-loss}
\mathcal{L}_{pre-train}(\theta) =\sum_{n=1}^{N} \log P_n[y_n].
\end{equation}

This pre-training stage is independent of the target style and it can be shared over different styles. After this stage, $f(\cdot)$ is able to generate realistic low-level details (which are missing in $I'$). 
In the next, we describe how a specific style  is incorporated in $f(\cdot)$ using a fine-tuning phase.

\subsection{Style-specific fine-tuning}
\label{sec.fine-tuning}

Given a style description provided with a textual sentence $t_s$, the goal is to fine-tune the pre-trained translator $f(\cdot)$ (Sec.~\ref{sec.upscaling}) to make it  generate image details in the style  of $t_s$.
We fine-tune only the decoder of $f(\cdot)$, keeping frozen the encoder.
% As the fine-tuning guidance we use the CLIP space and we maximize the cosine similarity between the image generated by $f(\cdot)$ and the projection of $t_s$  into this space. 
% However, merely maximizing this similarity does not take into account content preservation, so there is a risk of losing content information.
% However,  only maximizing this similarity can lead to completely destroy the original content contained in $I$ (Sec.~\ref{sec.Introduction}).
To ensure both stylization and content preservation, we design a prompt that consists of two parts, i.e., a style description $t_s$ and an image caption $t_a$,  which describe the image {\em content} and can be obtained from a generic image captioning dataset.
% at fine-tuning time
 % we introduce additional language supervision, i.e., a textual description of the source image {\em content}, noting as $t_a$, which can be obtained from a generic image captioning dataset.
% We concatenate $t_s$ and $t_a$ to get a joint content-and-style description of the desired image $I^s$, obtaining a {\em prompt}~\cite{radford2021learning} sentence $t_q$. 
For instance, given $t_a =$  \textit{``A man's hand is adjusting his black tie"} and $t_s =$  \textit{``cartoon"},
we obtain $t_q =$ \textit{``a cartoon of a man's hand is adjusting his black tie}.
On the other hand, in order to represent the image generated by $f(\cdot)$, we first need to sample the distributions in $P$
(Sec.~\ref{sec.upscaling}), and we do so using multinomial sampling:

\begin{equation}
\label{eq.argmax-sampling}
\hat{Y}_n = Sampling( P_n[y] ),\: \:  \:  \forall n \in \{1, ..., N\}.
% \hat{Y}_n = \arg \max_{y \in \{1, ..., M\}} P_n[y] \: \:  \:  \forall n \in \{1, ..., N\}.
\end{equation}
The sampled sequence $\hat{Y}$ is reshaped to a $2k \times 2k$ grid and fed to the image detokenizer to get the final image $I^s = G(\hat{Y})$. Finally, using the  CLIP visual and textual encoders
 we compute the cosine similarity on the CLIP space:

\begin{equation}
\label{eq.CLIP-similarity}
r = Sim_{CLIP} (I^s, t_q). 
\end{equation}

However, directly using Eq.~\ref{eq.CLIP-similarity} as the fine-tuning objective function is not possible because 
Eq.~\ref{eq.argmax-sampling} is not differentiable. 
In addition, since there is no ground truth for the tokens of stylized images we use RL to encourage the model to explore the answers in the latent space of a pretrained vector-quantized model.
We use the 
 REINFORCE algorithm \cite{williams1992rl} that updates the parameters  of $f_{\theta}(\cdot)$ using the CLIP-based
 reward $r$, which could keep awarding the model for achieving better-stylized results until the model reaches a limit. This leads to the gradient estimate:
\begin{equation}
    \label{eq.rl_update}
    \nabla_{\theta|_d} \mathcal{L}_{fine-tune}(\theta|_d) = \sum_{n=1}^N r \nabla_{\theta|_d} \log P_n[y_n], %p (y_n | V', V)
\end{equation}
%= \mathbb{E}_{\pi}\left[ r_k(\cdot) \nabla_{\theta} \log \pi (z_k | t_k) \right],
where $\theta|_d$ indicates the parameters of the decoder only (we found fine-tuning both could lead to content loss).
By maximizing  Eq.~\ref{eq.rl_update} we encourage $f(\cdot)$ to generate images having both the content and the style of the  prompt  $t_q$.

With respect to the whole method, the reason we use down-sampled versions of the content image is that ``style" is commonly assumed to be involved in the low-level visual details, such as colors, texture, painting strokes, etc. $X'$, which in our formulation represents $I$ at a lower resolution, presumably keeps most of the content in $I$ discarding some details, this way facilitating the style translation process. 
Preliminary experiments (in Appendix~\ref{apd:ablation}) in which we fed the encoder with tokenized full-resolution images  led to  poor results, demonstrating that the different cardinality 
between the source and the target sequence  is an essential component in this translation process. 

\begin{figure*}[ht]
% \vspace{-0.3cm}
    \begin{center}
    \includegraphics[width=\linewidth]{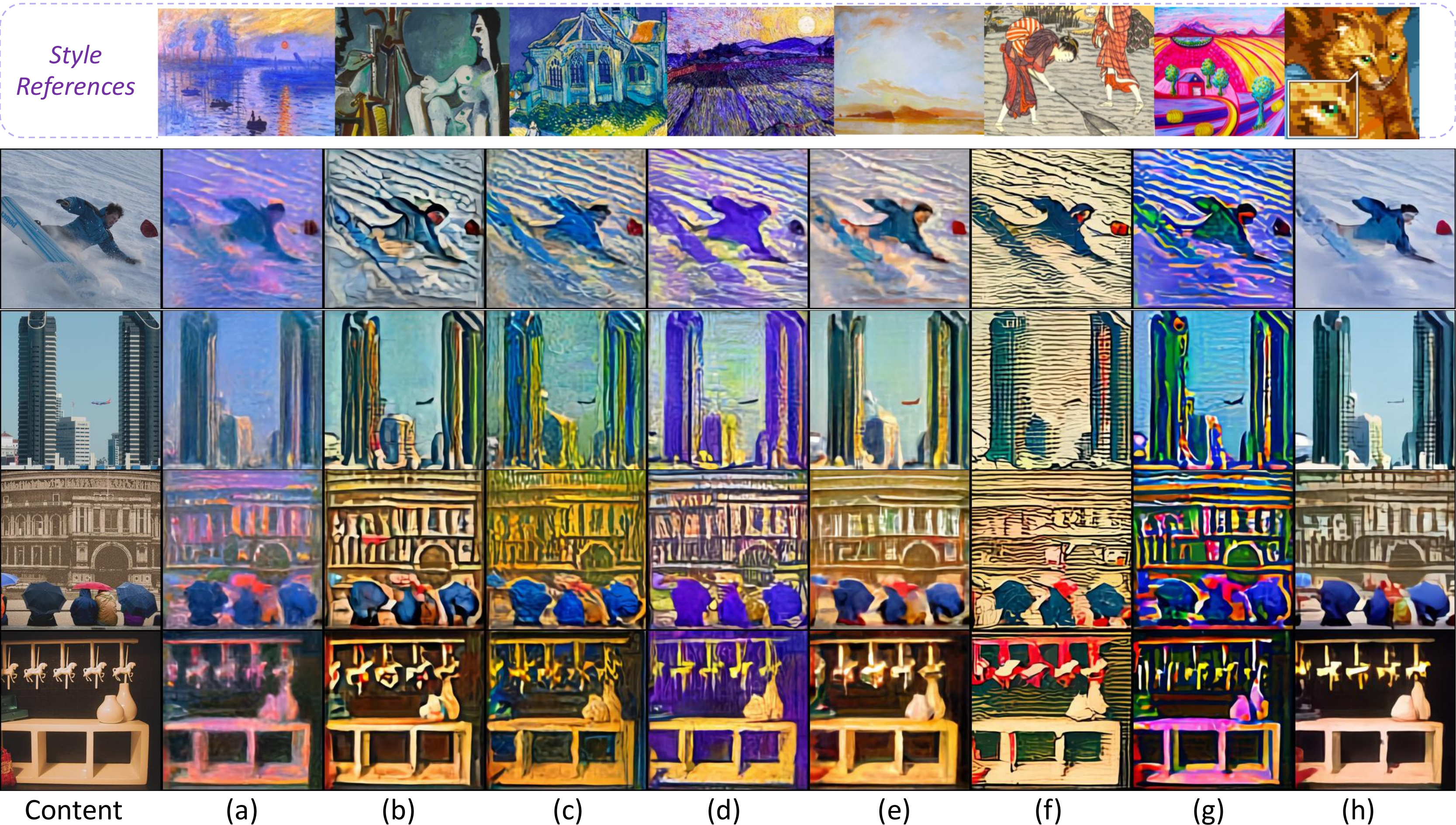}
    \end{center}
    % \vspace{-0.2cm}
    \caption{Qualitative results of $\mathtt{Styler}$DALLE-1 on various styles: (a) Monet Sunrise, (b) Picasso cubism, (c) Van Gogh blue color, (d) Van Gogh purple color, (e) warm and relaxing, (f) ukiyo-e print, (g) fauvism, (h) pixel art illustration. The style references at the top are for illustration only (not used as input to the model).}
    \label{fig:qla}
\end{figure*}

\begin{figure*}[ht]
    \begin{center}
    \includegraphics[width=\linewidth]{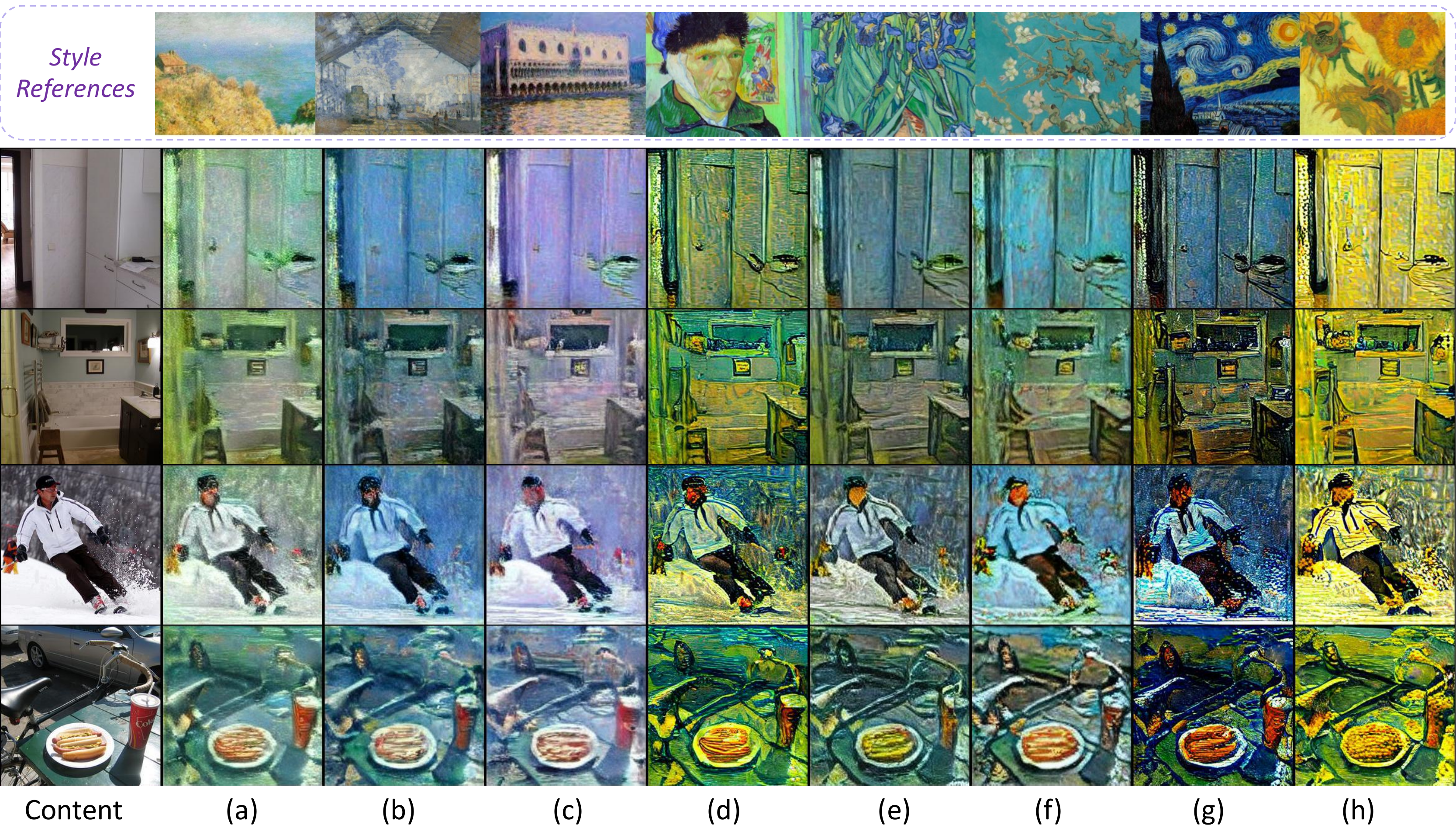}
    \end{center}
    % \vspace{-0.2cm}
    \caption{Qualitative results of $\mathtt{Styler}$DALLE-Ru on various styles: (a) Monet, (b) Monet Paris, (c) Monet Venice, (d) Van Gogh, (e) Van Gogh Irises, (f) Van Gogh Almond, (g) Van Gogh Starry Night, (h) Van Gogh Sunflowers. The style references at the top are for illustration only (not used as input to the model).}
    \label{fig:qlar}
\end{figure*}

\section{Experiments}

\subsection{Training Details}
We implement our method based on two types of pretrained vector-quantized tokenizer: 1) the officially released dVAE of DALL-E 1~\cite{dalle-dvae} and 2) the VQGAN of Ru-DALLE~\cite{ru-dalle}, and consequently obtain two groups of results, noting as $\mathtt{Styler}$DALLE-1 and $\mathtt{Styler}$DALLE-Ru, respectively. We use the two models in our experiment because they are open-sourced while our method is applicable to any large-scale pretrained vector-quantized tokenizer.

To train the model, we use the MS-COCO~\cite{lin2014microsoft} train-set, which contains 83k images of common objects in daily scenes. In the self-supervised pretraining stage, we only use the images while in the style-specific fine-tuning stage we use both images and captions.
The $\mathtt{Styler}$DALLE NAT model consists of a 4-layer encoder and an 8-layer decoder while the attention head number is 8 and the hidden dimension is 512.
We use Pytorch~\cite{NEURIPS2019_9015} to implement our method.
% To train our model, we use the train-set of COCO~\cite{lin2014microsoft} dataset, which contains 82,783 images while each image has 5 captions.
% In the self-supervised pre-training stage, we only use the images in the COCO train-set.
We train the NAT model for 25 epochs with a learning rate of 1e-4.
We use Adam~\cite{kingma2014adam} optimizer.
In the style-specific fine-tuning stage, we use both the images and the captions.
In particular, we utilize all the caption annotations to enhance the model robustness, as usually human-beings annotate different captions of a single image.
Notably, the caption annotations are only used at the fine-tuning stage. In other words, $\mathtt{Styler}$DALLE does not need to use image caption as input at inference time.
We only fine-tune the decoder of the NAT model, and keep the encoder frozen. We use Adam optimizer with a learning rate of 1e-6.
For CLIP model, we use the CLIP ViT-B/32 model.
For both training stages, the model is trained on a single RTX-A6000 GPU for 24 hours.
% We give more implementation details in Appendix~\ref{apd:details}.

\subsection{Experimental Results}
In the following, we present qualitative, quantitative, and user study results of our method and comparative methods, as well as comparisons with reference image-based methods.
% In this section, we show the stylized results of $\mathtt{Styler}$DALLE and a comparison with the results of two types of style transfer methods, i.e.,  language-guided methods and  standard reference image-based methods.
The additional implementation details, ablation study, inference time comparison, and additional experimental results are shown in Appendix ~\ref{apd:details}, ~\ref{apd:ablation}, ~\ref{apd:infert}, and ~\ref{apd:results}, respectively.

\begin{figure*}[ht]
    \begin{center}
    \includegraphics[width=\linewidth]{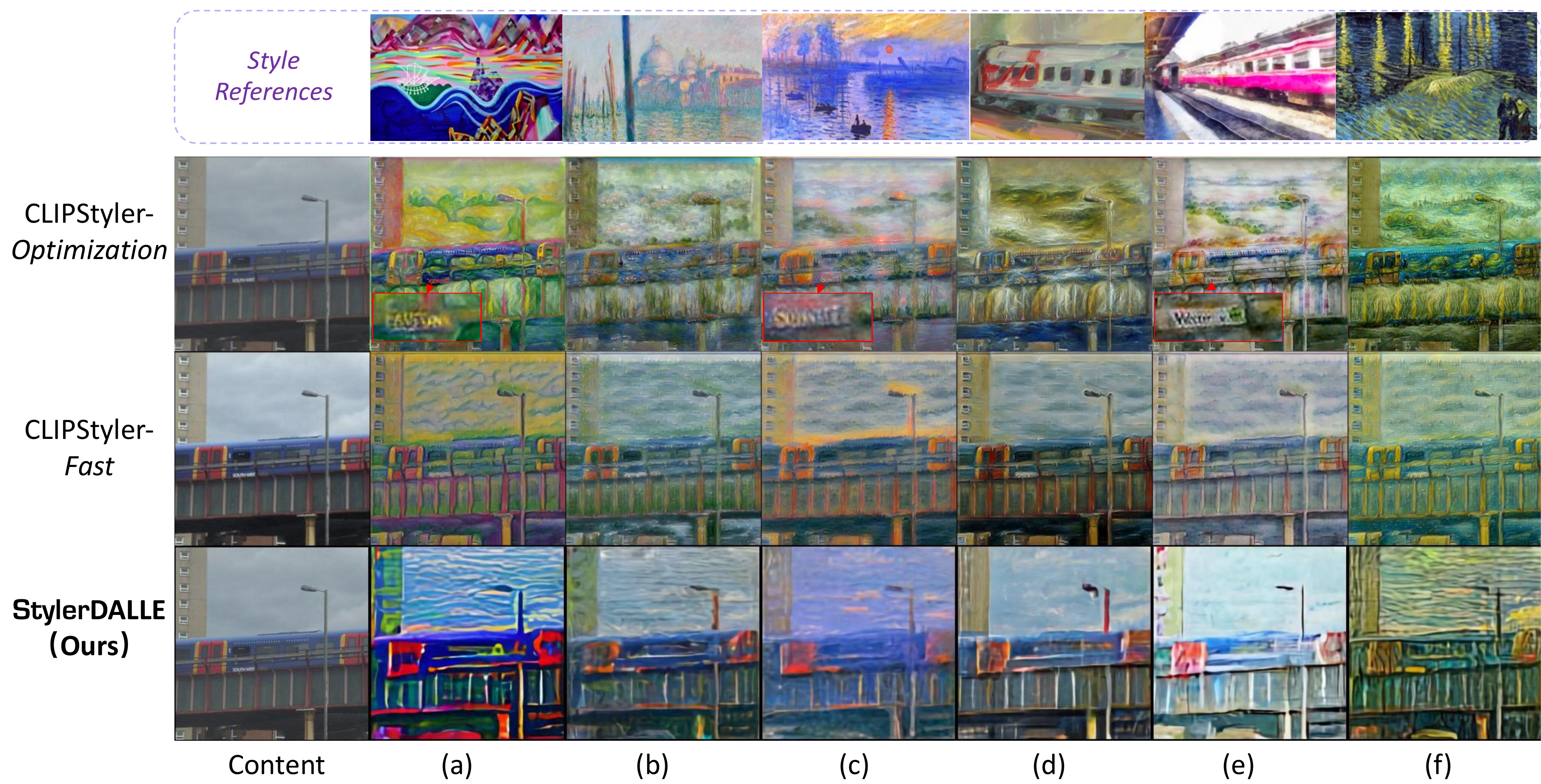}
    \end{center}
    % \vspace{-0.2cm}
    \caption{Comparisons with language-guided methods. The styles are (a) fauvism, (b) Monet, (c) Monet Impression Sunrise, (d) oil painting, (e) watercolor painting, (f) Van Gogh, (g) Van Gogh blue color. The style references at the top are for illustration only (not used as input to the model).}
    \label{fig:crl}
\end{figure*}

\begin{table*}[ht]
% \vspace{-0.3cm}
\centering
\begin{tabular}{c|c|c|c|c|c|c|c|c}
\toprule
\multicolumn{1}{c|}{Dataset} &
\diagbox{Method}{Style} &
  Fauvism &
  Monet &
   \begin{tabular}[c]{@{}c@{}}Monet\\ Sunrise\end{tabular} &
   \begin{tabular}[c]{@{}c@{}}Monet\\ Sunset\end{tabular} &
   \begin{tabular}[c]{@{}c@{}}Monet \\ Paris\end{tabular}&
  % \begin{tabular}[c]{@{}c@{}}Monet \\ Venice\end{tabular}&
  \begin{tabular}[c]{@{}c@{}}Van Gogh\\ Irises \end{tabular} &
    \begin{tabular}[c]{@{}c@{}}Van Gogh\\ Almond Blossoms \end{tabular}
  \\ \midrule
\multirow{2}{*}{CoCo} &
CLIPstyler-Fast &
  26.11 &
  27.46 &
   24.78 &
  26.70 &
    27.60 &
  26.23 &
  27.05 \\ \cline{2-9} 
&
 Ours &
  \bf 30.61 &
  \bf 27.72 &
  \bf 28.81 &
  \bf 30.68 &
    \bf 29.63 &
  \bf 28.14 & 
  \bf 31.64 \\ \midrule
  
\multirow{2}{*}{AFHQ} &
CLIPstyler-Fast &
  26.52 &
  26.08 &
   23.27 &
  25.05 &
    25.42 &
  23.28 &
  24.31 \\ \cline{2-9} 
&
 Ours &
  \bf 29.96 &
  \bf 26.55 &
  \bf 26.58 &
  \bf 29.11 &
    \bf 26.10 &
  \bf 28.04 & 
  \bf 31.92 \\ \midrule

\multirow{2}{*}{ImageNet-100} &
CLIPstyler-Fast &
  26.63 &
  27.16 &
   24.33 &
  26.12 &
    27.28 &
  26.32 &
  26.23 \\ \cline{2-9} 
&
 Ours &
  \bf 30.22 &
  \bf 27.53 &
  \bf 28.29 &
  \bf 30.28 &
    \bf 27.67 &
  \bf 28.26 & 
  \bf 31.60 \\
  
\bottomrule
\end{tabular}
\caption{Quantitative comparisons with CLIPstyler~\cite{kwon2022clipstyler} on different types of styles and datasets.}
\label{tab:qat}
% \vspace{-0.3cm}
\end{table*}

% \vspace{0.2cm}
\paragraph{Qualitative Results.}
In Fig.~\ref{fig:1}, Fig.~\ref{fig:qla} and Fig.~\ref{fig:qlar}, we show that our method can effectively transfer various types of styles, i.e., a) abstract art styles, e.g., ``fauvism" and ``pop art"; b) artist-specific styles, e.g., ``Monet" and ``Van Gogh"; c) artist-specific styles with additional descriptions, e.g., ``Monet Paris" and ``Van Gogh Sunflowers"; d) artistic painting types, e.g., ``pixel art illustration"; and e) emotional effects, e.g., ``warm and relaxing". 
According to the qualitative results, we draw the following conclusions:
1) $\mathtt{Styler}$DALLE can transfer abstract style concepts which go beyond the texture and color features and 
are similar to the typical trait of the artist/artistic target style;
2) each style corresponds to generated images that are different from those of other styles;
3) the image content is well preserved;
and 4) $\mathtt{Styler}$DALLE can be applied to  open-domain content images (i.e., the image content can contain animals, human beings, daily objects, buildings, etc.).

Further, we compare $\mathtt{Styler}$DALLE with the recent language-guided style transfer method CLIPStyler~\cite{kwon2022clipstyler}.
CLIPstyler proposes two methods:  1) CLIPStyler-Optimization, which optimizes a style transfer network {\em for each content image}, thus being time-consuming; and 2) CLIPStyler-Fast, which is the most comparable method to ours as it trains a network {\em for each style}, and then it can be used with any content images.
% , which is . Both $\mathtt{Styler}$DALLE and CLIPStyler can be applied to  open-domain content images and use only language instructions to describe the target style.
As shown in Fig.~\ref{fig:crl}, CLIPStyler-Optimization generates diverse stylized results but it suffers from inharmonious artifacts.
For instance, in the column ``Monet", there are many plants on the train.
% Other artifacts  are generated with the other styles, e.g., ``fauvism", ``Monet" and ``Van Gogh".
In addition, CLIPStyler-Optimization has the problems of writing the style text in the results, as in the results of "fauvism" and so on.
On the other hand,
the images generated by CLIPStyler-Fast do not contain artifacts but there is  less variation among different styles.
Importantly, it is hard to recognize the  typical trait  of each artistic style, and the main differences among the styles are the colors.
In contrast, the results of $\mathtt{Styler}$DALLE are much closer to the artworks of the specific artistic style, they show distinct differences among different styles, and they have no artifact issues.
% Moreover, $\mathtt{Styler}$DALLE is much less time-consuming as compared to CLIPStyler-Optimization. 
% A magnified comparison is illustrated in the appendix~\ref{apd:results}.

% \vspace{0.2cm}
\paragraph{Quantitative Results.}
To do quantitative analysis, we use the CLIP similarity score, formalized as $score = Sim_{CLIP} (I, t_s)$, which is computed between generated stylized images and textual description of the target style.
In Tab.~\ref{tab:qat}, we present the results of $\mathtt{Styler}$DALLE-1 and CLIPstyler-Fast.
Since both methods are applicable for arbitrary content images, we use the MS-COCO val-set, AFHQ val-set~\cite{afhq}, and ImageNet-100 val-set~\cite{im100} for evaluation.
According to the results on multiple datasets, although the way we use CLIP is to compute rewards in Reinforcement Learning, instead of directly using CLIP-scores to optimize the network as CLIPstyler, we achieve comparable and even better quantitative results, indicating the effectiveness of our method.

% \vspace{0.2cm}
\paragraph{User Study.}
We conduct a user study to see human opinions towards the stylized images coming from different methods: 1) CLIPstyler-Fast, 2) $\mathtt{Styler}$DALLE-1, and 3) $\mathtt{Styler}$DALLE-Ru.
In specific, we collect opinions from 35 human subjects with a 30-question questionnaire. In each question, we ask them to select one stylized result that is closest to a target style.
As in Tab.~\ref{tab:us}, among the three methods, $\mathtt{Styler}$DALLE-Ru achieves the highest preference score of 68.76\%, indicating the superiority of our method.
We also find that humans prefer $\mathtt{Styler}$DALLE-Ru much more than $\mathtt{Styler}$DALLE-1.
We infer this could be because the results from $\mathtt{Styler}$DALLE-1 are blurry, as they are  based on the dVAE of DALLE-1, and humans dislike blurry images.
More user study details are given in \ref{apd:usd}.
\begin{table}[t]
\centering
% \vspace{-0.2cm}
\begin{tabular}{c|c|c|c}
\toprule
(\%) &
  \begin{tabular}[c]{@{}c@{}}CLIPstyler-\\Fast\end{tabular} &
   \begin{tabular}[c]{@{}c@{}}$\mathtt{Styler}$\\DALLE-1\end{tabular} &
  \begin{tabular}[c]{@{}c@{}}$\mathtt{Styler}$\\DALLE-Ru \end{tabular}\\ \midrule
Preference &
  15.90 &
  15.33 &
  \bf 68.76 \\ \bottomrule
\end{tabular}
\caption{Preference scores of user study.}
\label{tab:us}
% \vspace{-0.3cm}
\end{table}

\begin{figure*}[ht]
    \begin{center}
    \includegraphics[width=\linewidth]{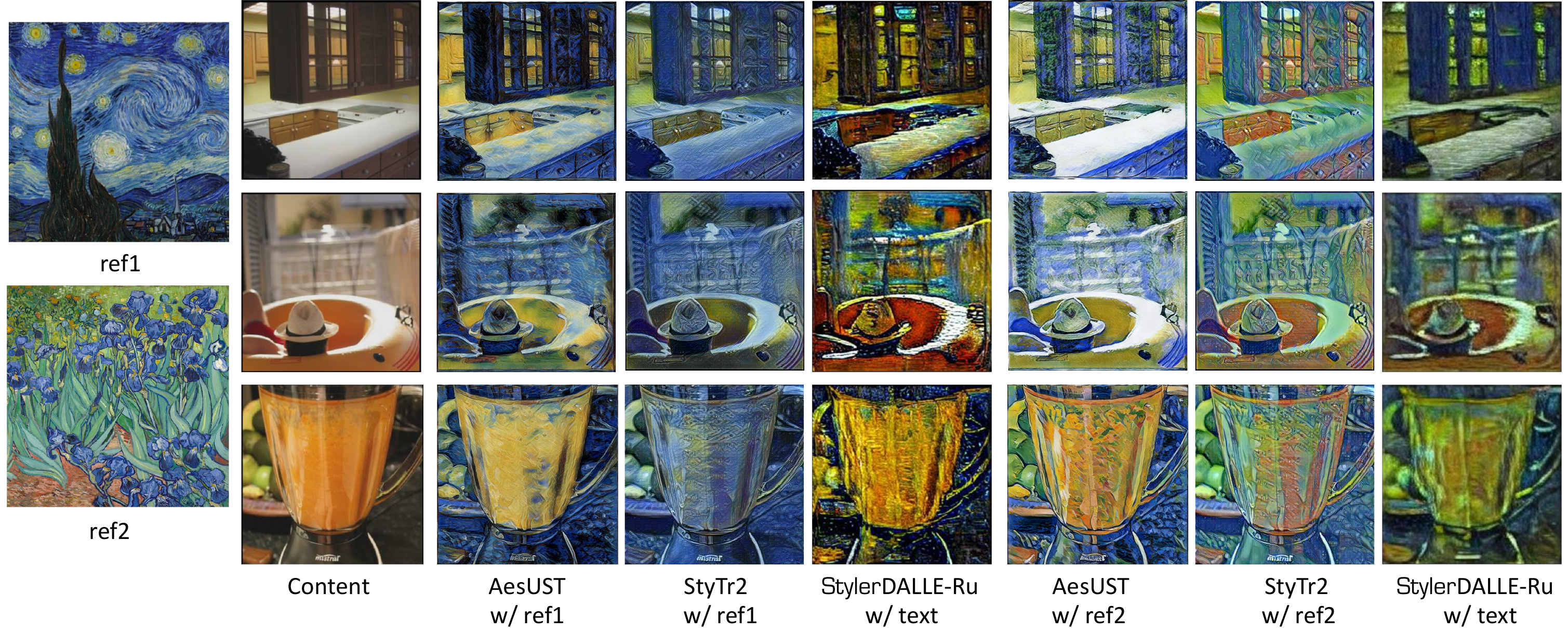}
    \end{center}
    % \vspace{-0.2cm}
    \caption{Comparisons with reference image-based methods. $\mathtt{Styler}$DALLE is able to transfer more abstract concepts, e.g., specific painting strokes, and is less likely to produce semantic errors.}
    \label{fig:crr}
    % \vspace{-0.1cm}
    \end{figure*}

% \vspace{0.2cm}
\paragraph{Comparisons with Reference Image-Based Methods.}
We   compare $\mathtt{Styler}$DALLE with state-of-the-art reference image-based methods: 1) AesUST~\cite{wang2022aesust}, an arbitrary style transfer method which enhances the aesthetic reality using a GAN trained with a collection of artworks; and 2) StyTr2~\cite{deng2022stytr2}, an arbitrary style transfer method which uses a transformer to eliminate the biased content representation issues of CNN-based methods.
To make the comparison feasible, we show the results of AesUST and StyleTr2 using two Van Gogh paintings as the reference images, and the results of $\mathtt{Styler}$DALLE trained using the corresponding language description.

As shown in Fig.~\ref{fig:crr}, the results of $\mathtt{Styler}$DALLE are distinctive from the results of reference image-based methods.
In concrete, the results of both AesUST and StyleTr2 are mostly affected by the colors and the textures of the reference images, and to some extent in an unnatural way.
For instance,
on the bottom line, in the ``Van Gogh Irises" stylized results of AesUST and StyleTr2, the textures of the irises get transferred to the orange cup.
In transferring the style of ``Van Gogh Starry Night", the objects in the results of AesUST and StyleTr2 are mostly in the same dark-sky blue as the reference image, making the scenes a bit in-realistic.
By contrast, the colors and textures used in the reference style are well transferred while being appropriately applied to the contents, without changing the original semantics.
Moreover, $\mathtt{Styler}$DALLE achieves to transfer higher-level style features, e.g., the brushstroke, rather than merely colors and textures, leading to the results of a similar style to the target one.
For example, the strokes in {\em Starry Night} are sharper as compared to the ones in {\em Irises}, and the differences are also presented in the stylized results of  $\mathtt{Styler}$DALLE.

\section{Conclusion}
We present $\mathtt{Styler}$DALLE, a language-guided style transfer method that leverages the power of the large-scale pretrained vector-quantized image tokenizer and CLIP. Specifically, inspired by natural language translation, we propose a non-autoregressive sequence translation approach to manipulate the discrete visual tokens, from the content image to the stylized image. We use Reinforcement Learning to include CLIP-based language supervision on the style and content. 
Differently from previous work, $\mathtt{Styler}$DALLE can transfer abstract style concepts that are implicitly represented in the pretrained image tokenizer and CLIP and which cannot be easily obtained using reference images. Moreover, using the large-scale pretrained latent space as the basic image representation makes it possible to reduce the artifacts and the semantic incoherence better than the previous work that operates at the pixel level.

\paragraph{Acknowledgment.}
This work was supported by the MUR PNRR project FAIR (PE00000013) funded
by the NextGenerationEU and by the PRIN project CREATIVE (Prot. 2020ZSL9F9).

{\small
\bibliographystyle{ieee_fullname}
\bibliography{egbib}
}

\clearpage
\appendix
\section{Appendix}

\subsection{Implementation Details}
\label{apd:details}
For pretrained vector-quantized image tokenizer, we use the officially released dVAE of DALL-E\footnote{DALL-E: \url{https://github.com/openai/dall-e}} or the VQGAN of Ru-DALLE\footnote{Ru-DALLE:\url{https://github.com/ai-forever/ru-dalle}}.

To compare with CLIPStyler, we use the official implementation.\footnote{CLIPStyler: \url{https://github.com/cyclomon/CLIPstyler}}
For all reference image-based comparing methods, we use the officially released trained models.\footnote{ AesUST:~\url{https://github.com/EndyWon/AesUST}, StyTr2:~\url{https://github.com/diyiiyiii/StyTR-2}.}

\subsection{Ablation Study}
\label{apd:ablation}
We study two ablations of $\mathtt{Styler}$DALLE: (1) without captions, and (2) without scaling. 

\begin{figure*}[ht]
  \centering
  \subfigure[Ablation study results on $\mathtt{Styler}$DALLE-1, which is implemented based on the officially released dVAE of DALL-E~\cite{dalle-dvae}.]{
  \begin{minipage}[]{\linewidth}
  \centering
    \includegraphics[width=.7\linewidth]{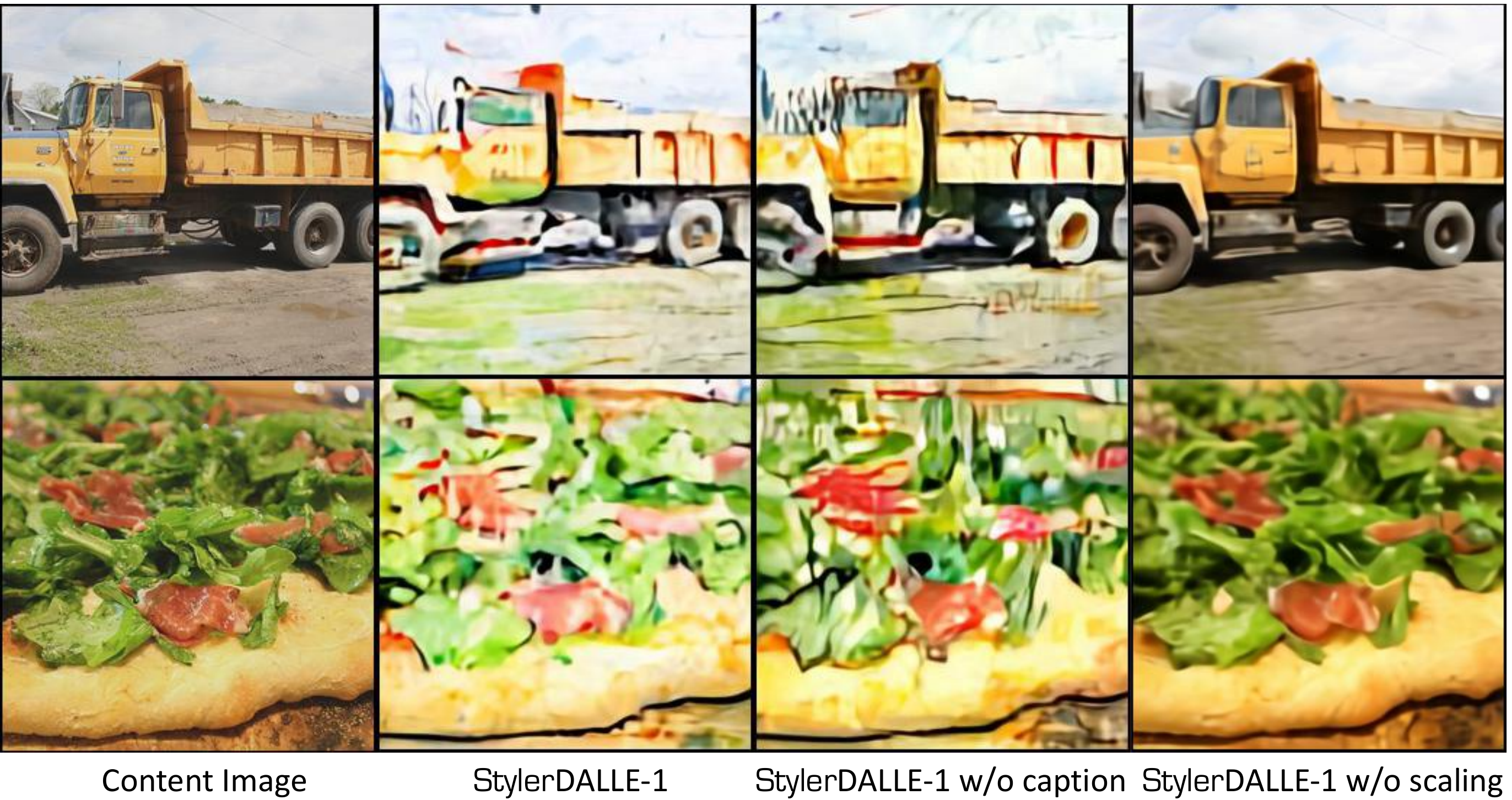}
    \label{fig:alb-1}
\end{minipage}}
% \vspace{-0.2cm}
\subfigure[Ablation study results on $\mathtt{Styler}$DALLE-Ru, which is implemented based on the VQGAN of Ru-DALLE~\cite{ru-dalle}.]{
\begin{minipage}[]{\linewidth}
\centering
    \includegraphics[width=.7\linewidth]{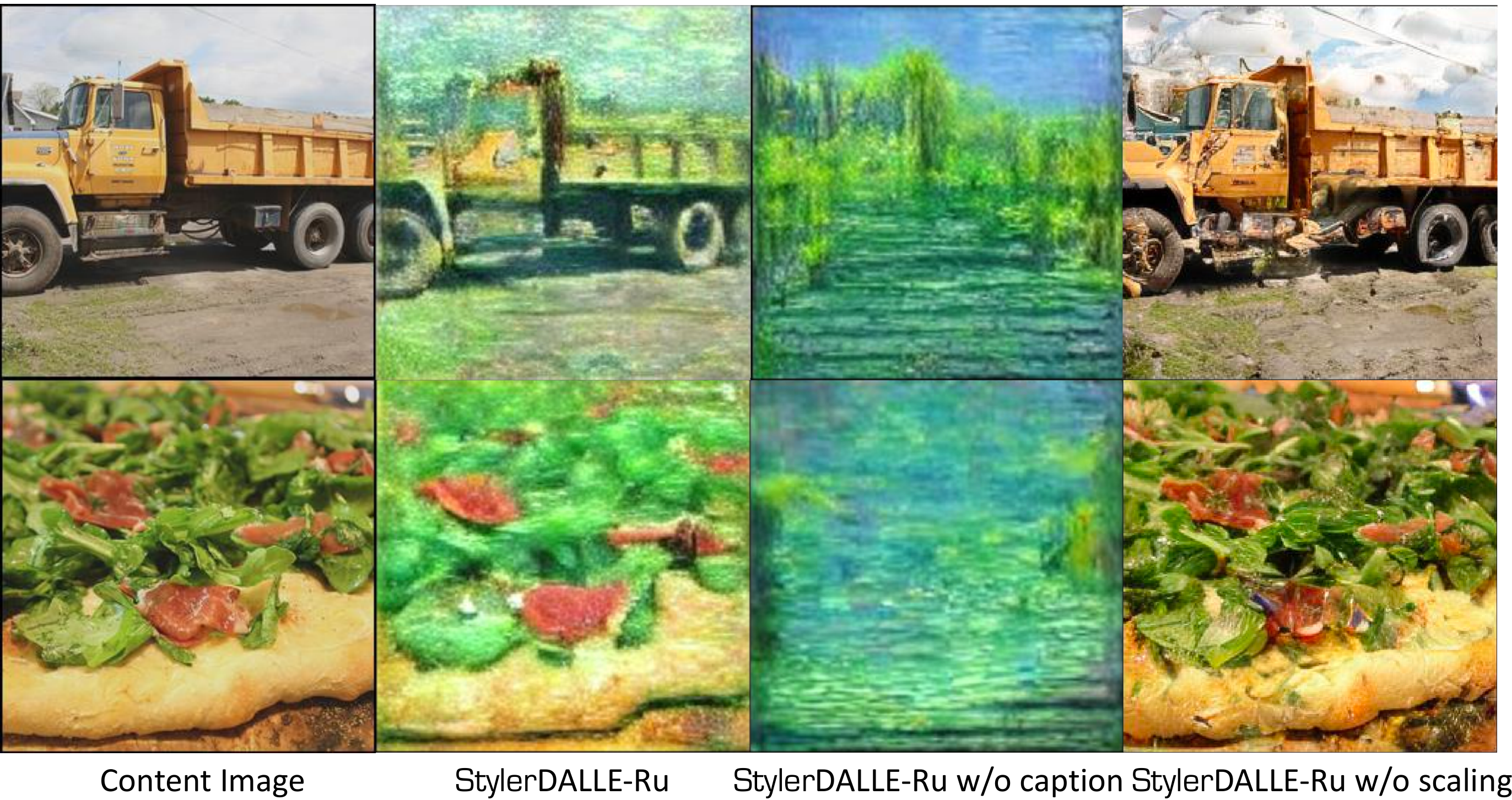}
    \label{fig:alb-ru}
  \end{minipage}}
  \caption{Ablation study on $\mathtt{Styler}$DALLE.}
  \label{fig:ablation}
  % \vspace{-0.3cm}
\end{figure*}

Firstly,  we ablate the usage of captions in formulating the prompt-based reward during the style-specific fine-tuning stage (Sec.~\ref{sec.fine-tuning}).
In more detail, instead of using the CLIP similarity between the stylized image $I^s$ and the prompt $t_q$ (which combines the style description $t_s$ and the image caption $t_a$) as the reward, we discard  $t_a$ and we compute the CLIP similarity between the stylized image $I^s$ and the style description $t_s$ as the reward.
As shown in Fig.~\ref{fig:ablation}, the two models $\mathtt{Styler}$DALLE-1 and $\mathtt{Styler}$DALLE-Ru show different results on the ablation ``w/o captions".
For $\mathtt{Styler}$DALLE-1 (Fig.~\ref{fig:alb-1}), we see that the results of the full model are slightly better. In the results of $\mathtt{Styler}$DALLE-1, the details are preserved better, and the colors are closer to the light and muted colors used in watercolor painting. Moreover, the results are overall harmonious as there are few abrupt brushstrokes.
Meanwhile, ``$\mathtt{Styler}$DALLE-1 w/o captions" also presents a satisfying style transfer quality, as the results keep a good balance between the stylization and content maintainess.
This indicates our method can also work for the dVAE of DALL-E when no caption  is provided, thus being less annotation-dependent.
Nevertheless, ``$\mathtt{Styler}$DALLE-Ru  w/o captions" (Fig.~\ref{fig:alb-ru}) fails to keep content consistency, emphasizing the significance of using captions as part of the language supervision in the Reinforcement Learning process for maintaining the content.
% $\mathtt{Styler}$DALLE-1 is not affected much by not using the captions in formulating the language supervision in Reinforcement Learning, while $\mathtt{Styler}$DALLE-Ru fails to keep content consistency.

Secondly, we ablate the operation  of  down-sampling  as introduced in Sec.~\ref{sec.Method}.
Specifically, we directly input the discrete tokens of the full-resolution image to the NAT model while conducting the same self-supervised pre-training and style-specific fine-tuning.
As shown in the results of `` $\mathtt{Styler}$DALLE-1 w/o scaling" (Fig.~\ref{fig:alb-1}) and ` $\mathtt{Styler}$DALLE-Ru w/o scaling" (Fig.~\ref{fig:alb-ru}), scaling is an important procedure in $\mathtt{Styler}$DALLE:  when the NAT model is input with the discrete tokens of the full-resolution image, the style cannot be incorporated effectively through the Reinforcement Learning fine-tuning stage.

\subsection{Inference Time}
\label{apd:infert}
% To compute the inference time of $\mathtt{Styler}$DALLE, .
To generate a single 256$\times$256 stylized image (including the time to down-sample, encode, translate through the NAT, and decode), $\mathtt{Styler}$DALLE needs $0.076$s, which is the average time computed over the COCO val-set using an RTX-A6000 GPU.
The main idea of our paper is to use large-scale pretrained image generative models for style transfer and we focus on using vector-quantization-based methods.
Therefore, we conclude that as compared to style transfer methods that are based on large-scale diffusion models, $\mathtt{Styler}$DALLE has the advantage of having less inference time.
For instance, as reported in a recent paper~\cite{chang2023muse}, Imagen~\cite{saharia2022photorealistic} takes $9.1$s to generate a 256$\times$256 image on TPUv4 accelerators.

\subsection{User Study Details}
\label{apd:usd}
\begin{figure*}[ht]
  \centering
  \subfigure[We illustrate each style with several examples to let the participant know the styles.]{
  \begin{minipage}[]{\linewidth}
  \centering
    \includegraphics[width=.5\linewidth]{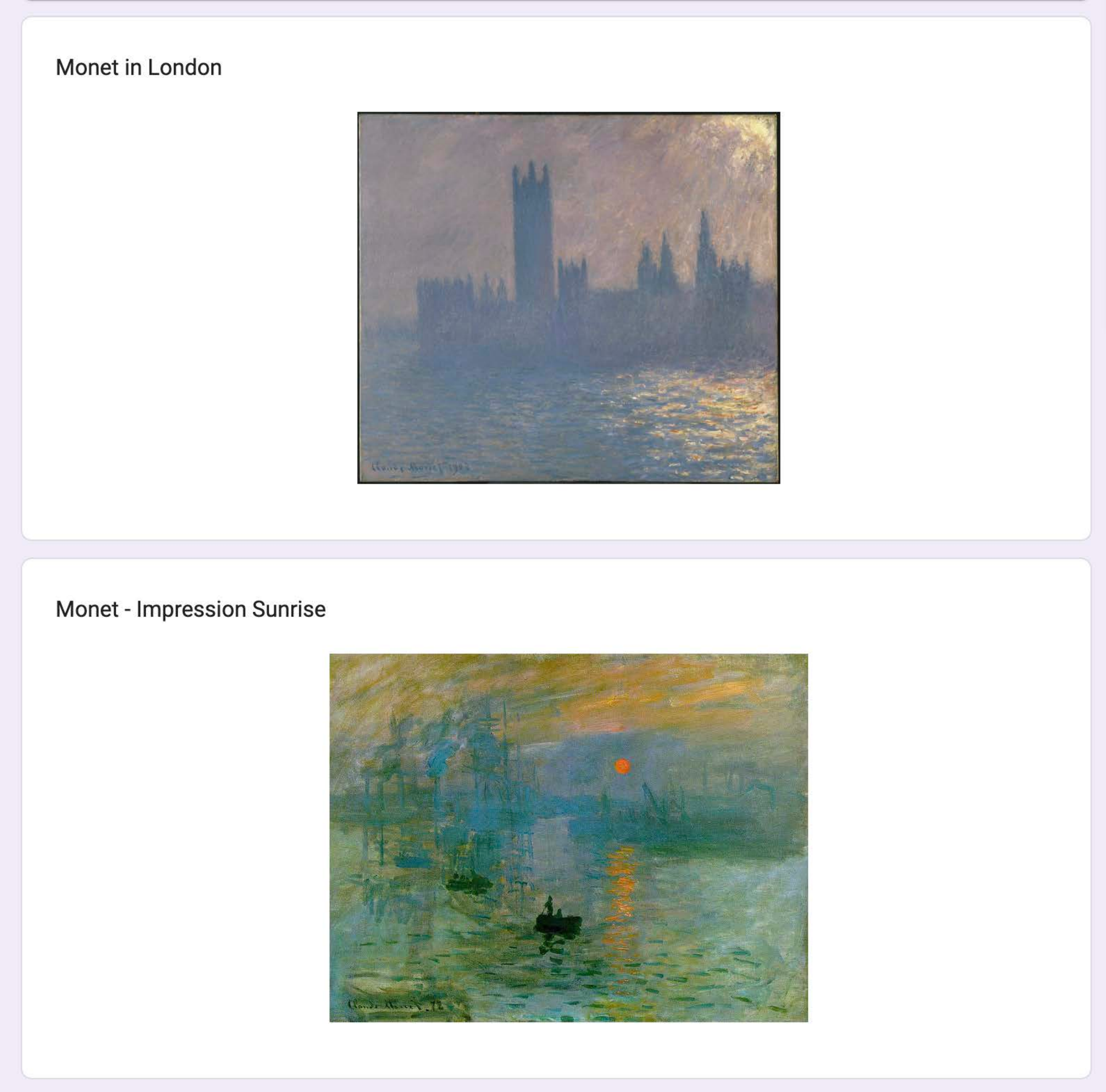}
    % \caption{``visionary art"}
    \label{fig:usd1}
\end{minipage}}
% \vspace{-0.2cm}
\subfigure[In each question, we ask the participant to select one image that is most likely to be of the target style. The order of the candidates is randomly shuffled.]{
\begin{minipage}[]{\linewidth}
\centering
    \includegraphics[width=.5\linewidth]{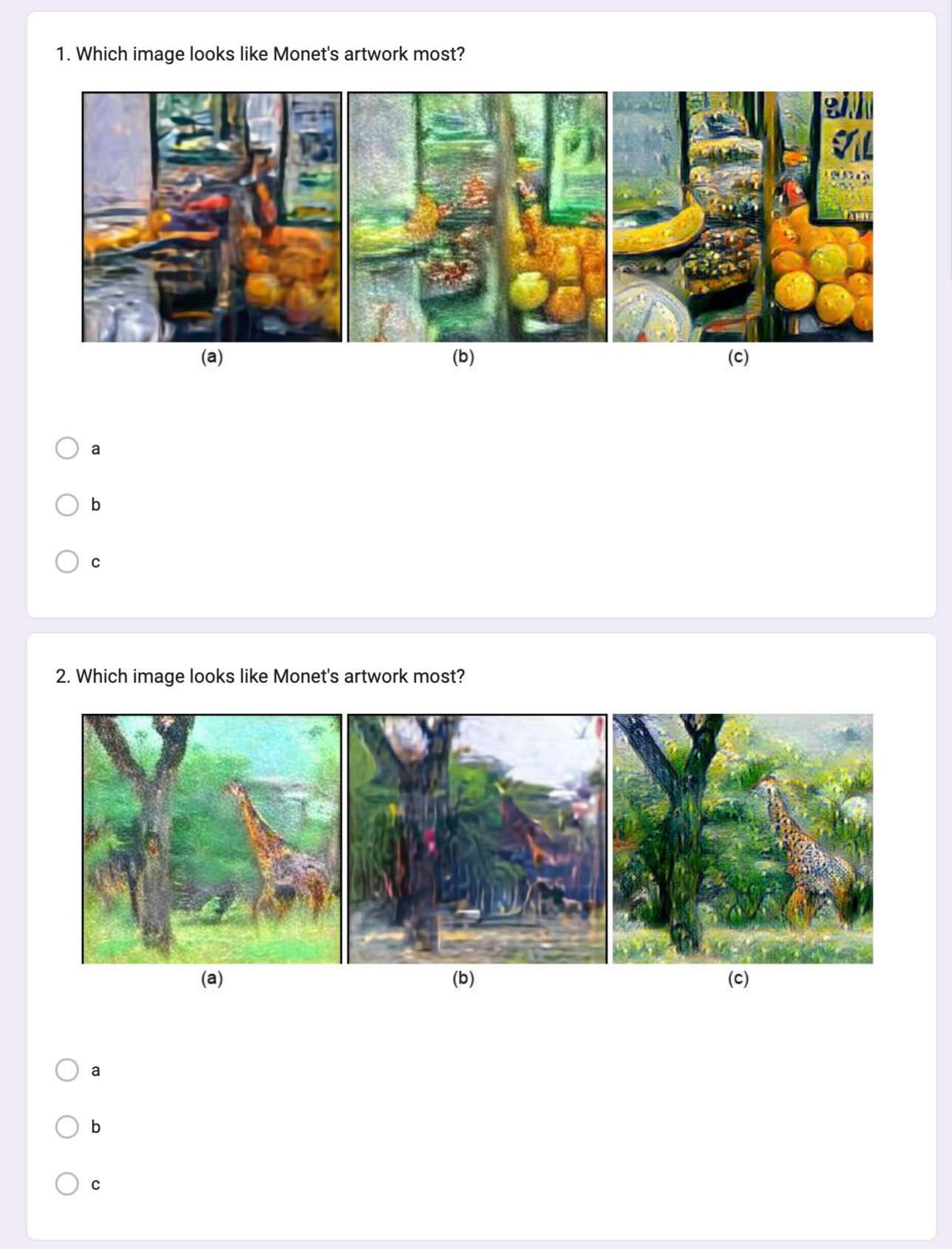}
    % \caption{``cartoon"}
    \label{fig:usd2}
  \end{minipage}}
  \caption{Illustrations of the user study details.}
  \label{fig:usd}
  % \vspace{-0.3cm}
\end{figure*}

Other than the quantitative analysis and qualitative analysis, as in Tab.~\ref{tab:us}, we further involve human subjects to evaluate the style transfer results of $\mathtt{Styler}$DALLE and the comparing method CLIPstyler.
To help the participants know the styles, at the beginning of evaluating each style, we incorporate several illustrations of the style (Tab.~\ref{fig:usd1}).
We show part of the questionnaire in Tab.~\ref{fig:usd2}.
We use Google Forms to collect user opinions.

\subsection{Additional Experimental Results}
\label{apd:results}

\begin{figure*}[ht]
    \begin{center}
    \includegraphics[width= .95\linewidth]{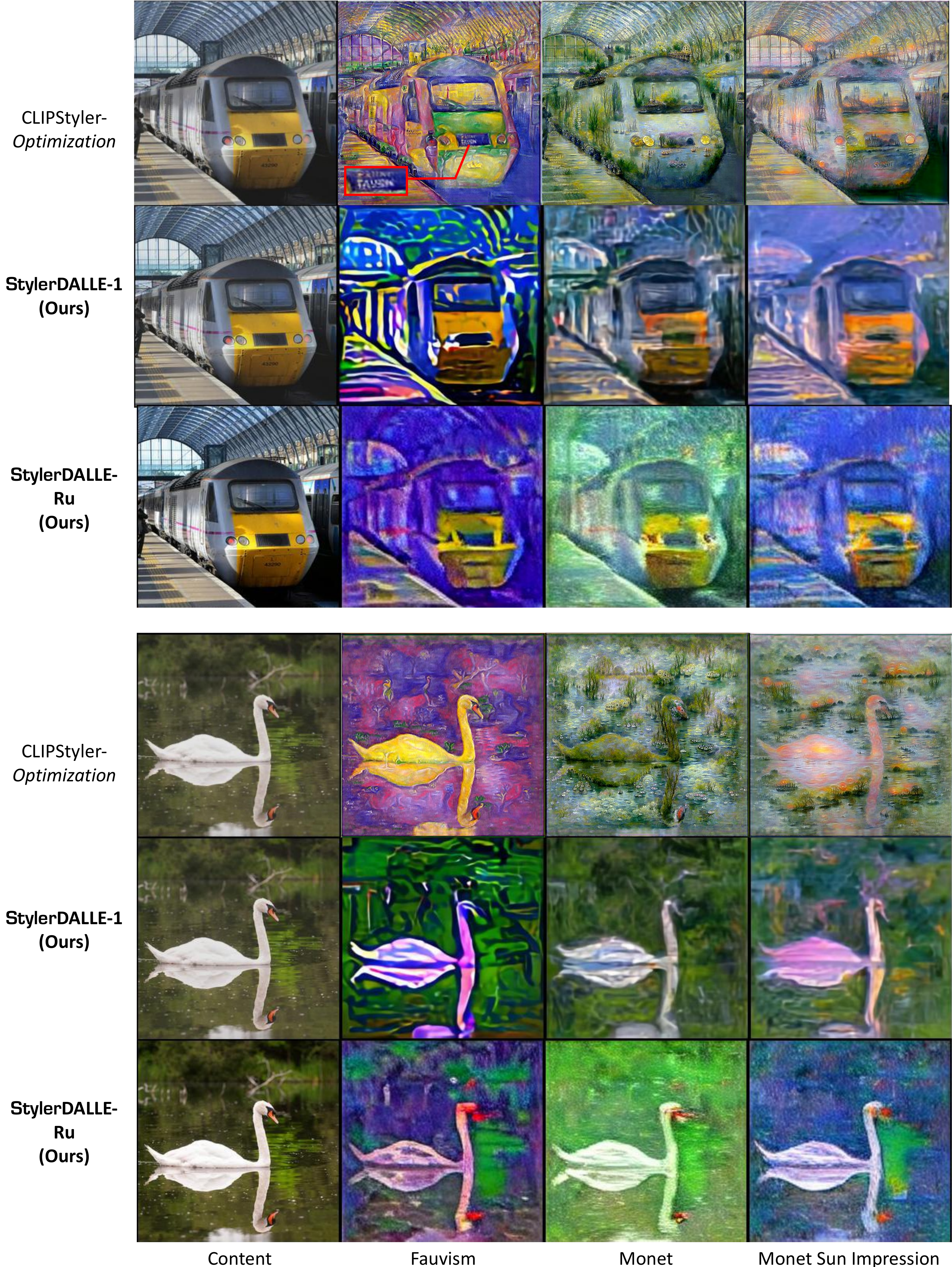}
    \end{center}
    \caption{Comparisons between $\mathtt{Styler}$DALLE and CLIPStyler, styles are shown on the bottom.}
    \label{fig:apd1}
    \end{figure*}

\paragraph{Additional Comparison Results.}
In Fig.~\ref{fig:apd1}, we illustrate the additional comparing results between $\mathtt{Styler}$DALLE and CLIPStyler-Optimization (i.e., the mainly proposed method in the paper). 
As shown, CLIPStyler-Optimization suffers from two issues. Firstly, there are many inharmonious artifacts that appear in the stylized images. For example, there are many plant-like artifacts in the stylized results of ``Monet" and multiple suns in the ``Monet Sun Impression" results. 
Secondly, the texts related to the language instructions are written in stylized images unexpectedly.
For instance, as in the top example of the ``fauvism" train, the written text ``fauvism" is on the front of the bus.

On the contrary, both $\mathtt{Styler}$DALLE-1 and $\mathtt{Styler}$DALLE-Ru do not have the above two issues. Furthermore, our results achieve well-characterized stylization results consistent with language instructions, and different styles are expressed with varied and distinctive brushstrokes related to the specific style.
In the following, we give more generated results of $\mathtt{Styler}$DALLE-1 and $\mathtt{Styler}$DALLE-Ru.

\begin{figure*}[ht]
    \begin{center}
    \includegraphics[width= .85\linewidth]{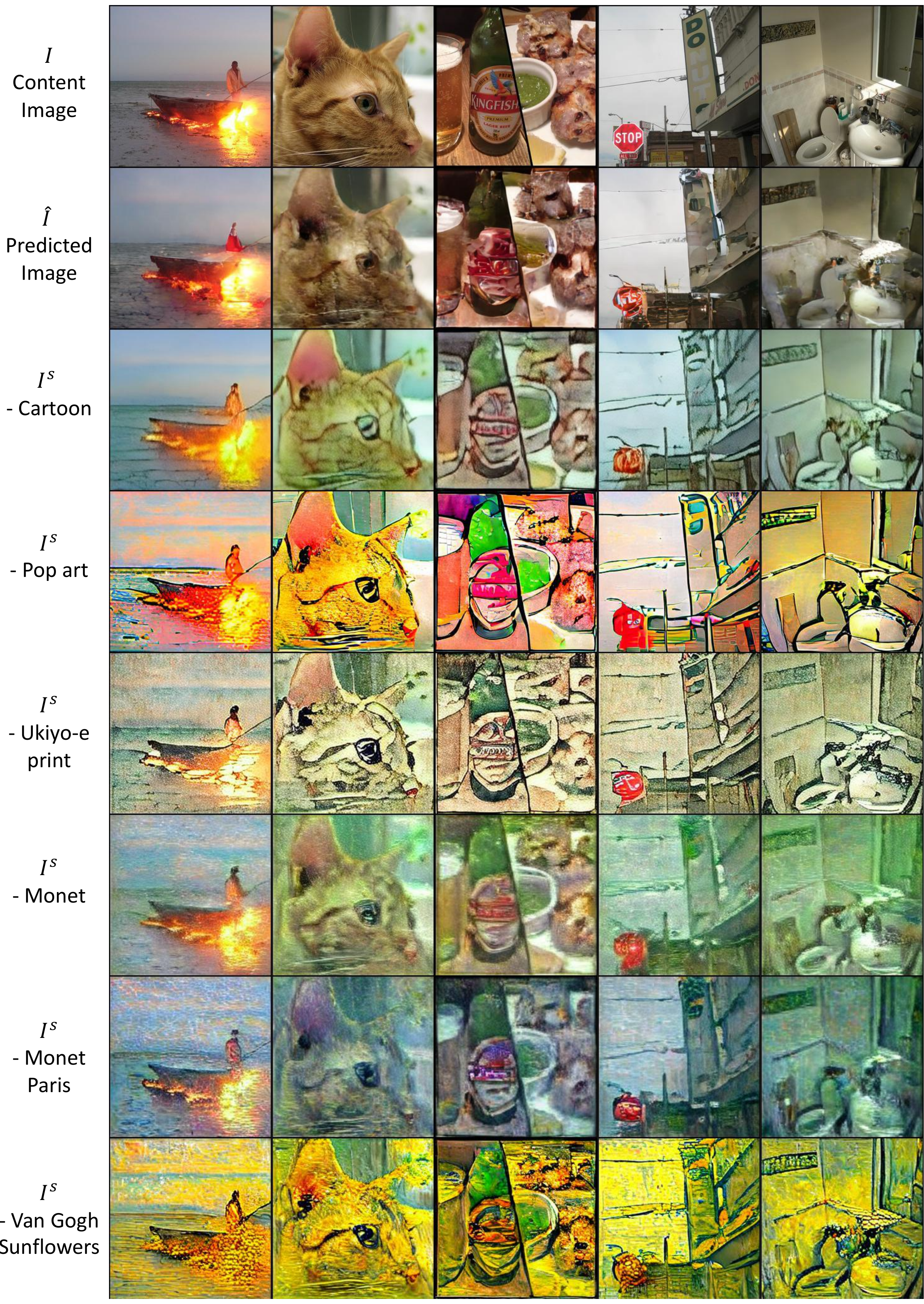}
    \end{center}
    \caption{Additional stylized results of $\mathtt{Styler}$DALLE-Ru.}
    \label{fig:apdru3}
\end{figure*}

\begin{figure*}[ht]
    \begin{center}
    \includegraphics[width= .85\linewidth]{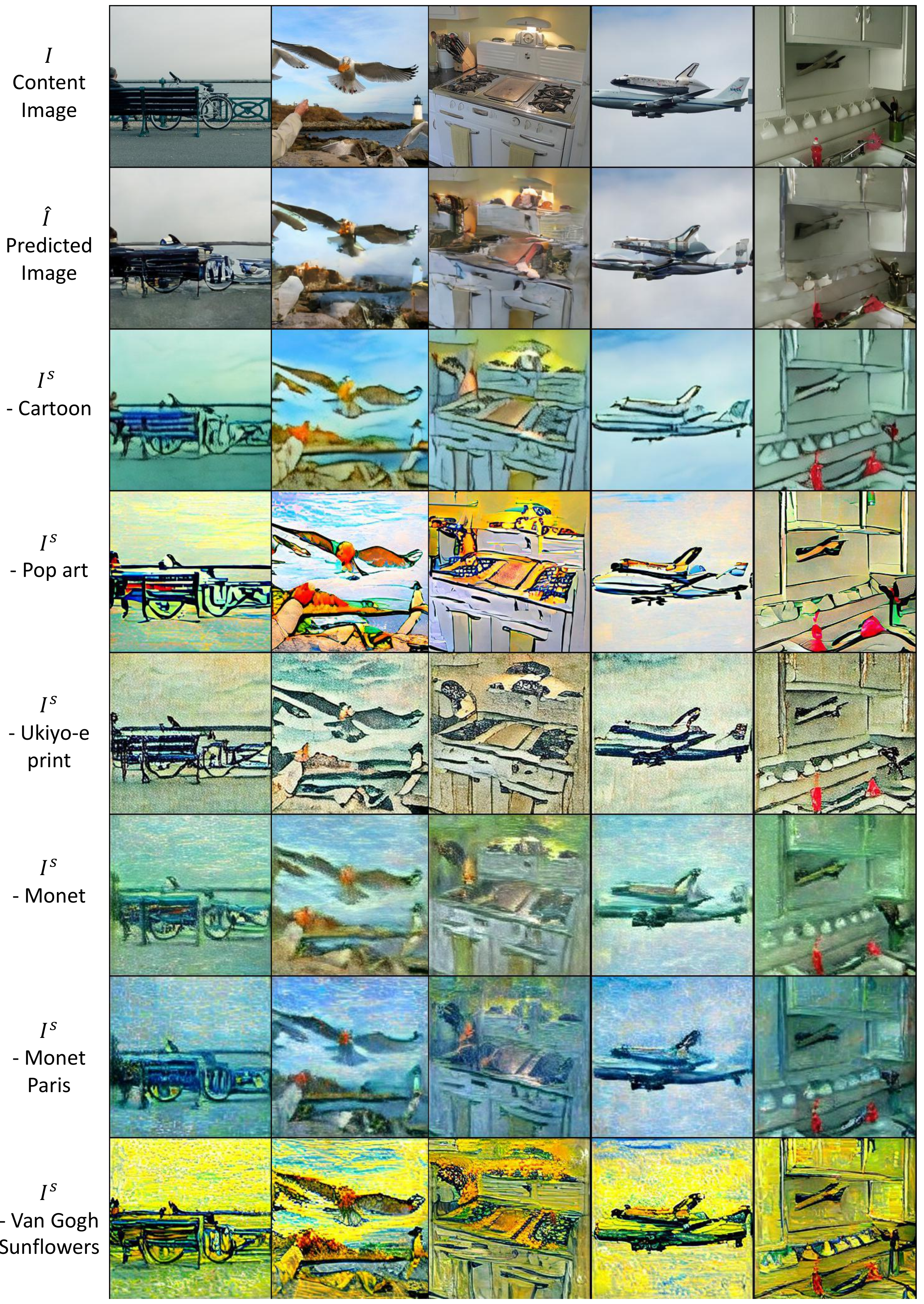}
    \end{center}
    \caption{Additional stylized results of $\mathtt{Styler}$DALLE-Ru.}
    \label{fig:apdru2}
\end{figure*}

\begin{figure*}[ht]
    \begin{center}
    \includegraphics[width= .85\linewidth]{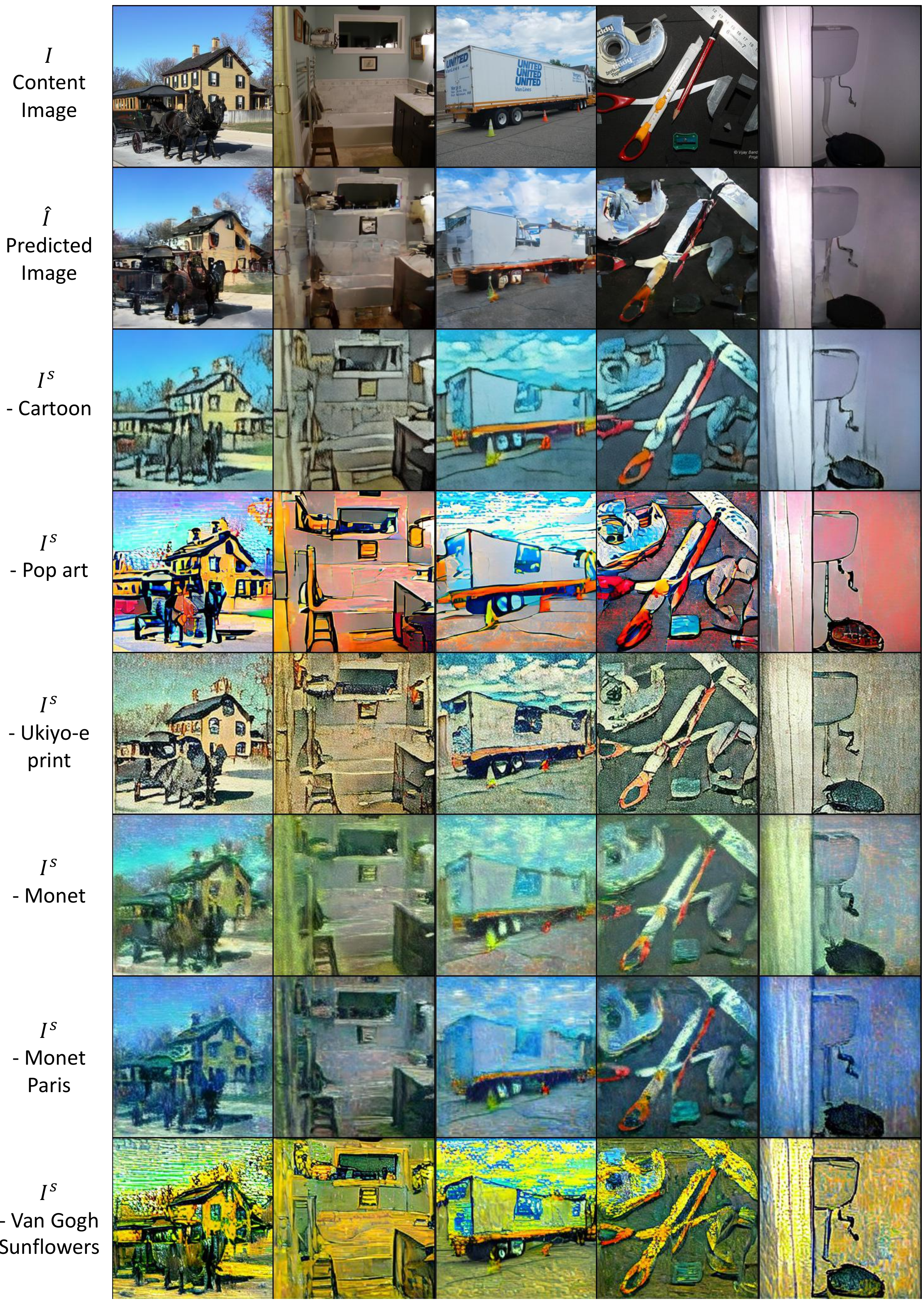}
    \end{center}
    \caption{Additional stylized results of $\mathtt{Styler}$DALLE-Ru.}
    \label{fig:apdru1}
\end{figure*}

\begin{figure*}[ht]
    \begin{center}
    \includegraphics[width= .85\linewidth]{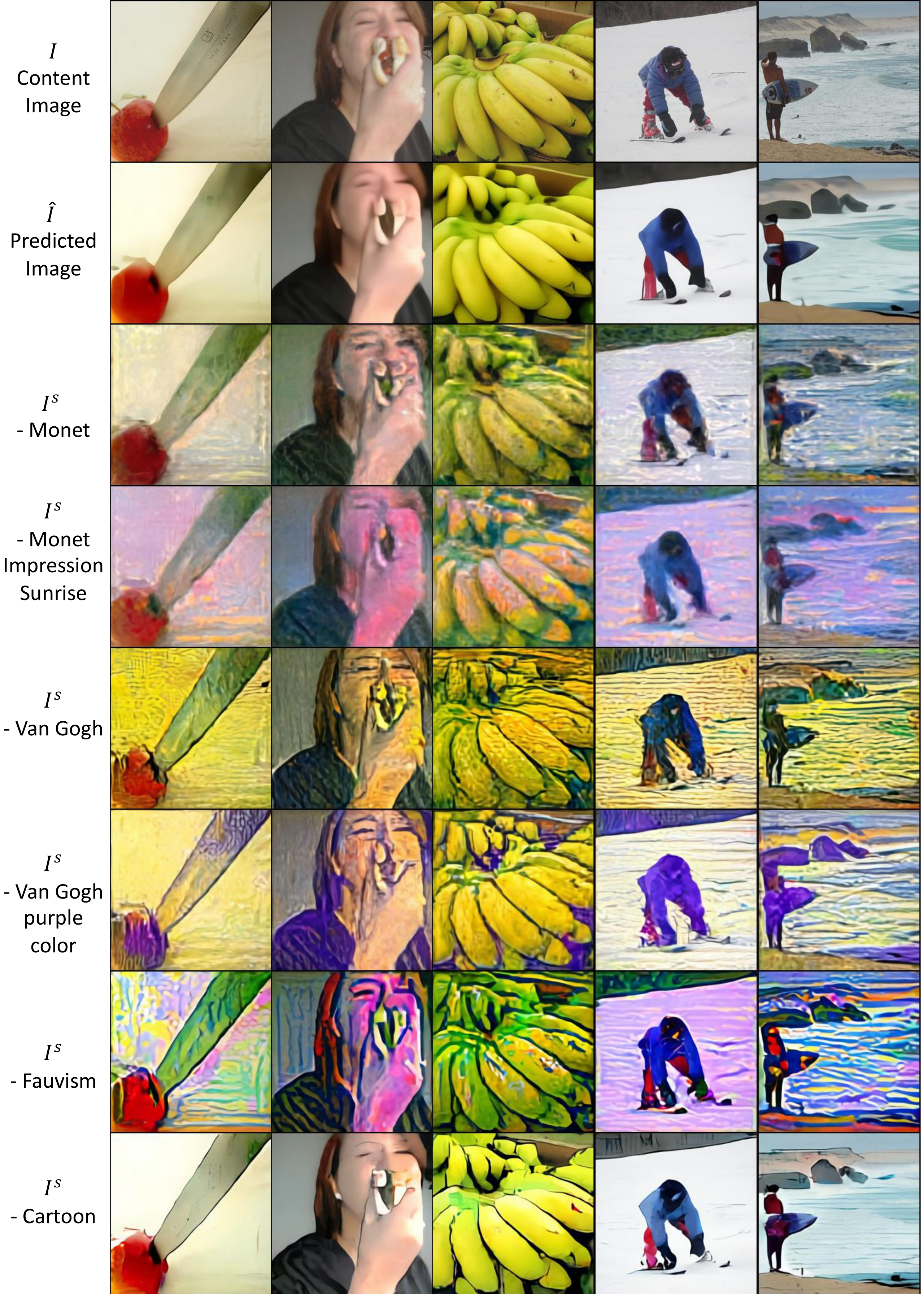}
    \end{center}
    \caption{Additional stylized results of $\mathtt{Styler}$DALLE-1.}
    \label{fig:apd2}
\end{figure*}

\begin{figure*}[ht]
    \begin{center}
    \includegraphics[width= .85\linewidth]{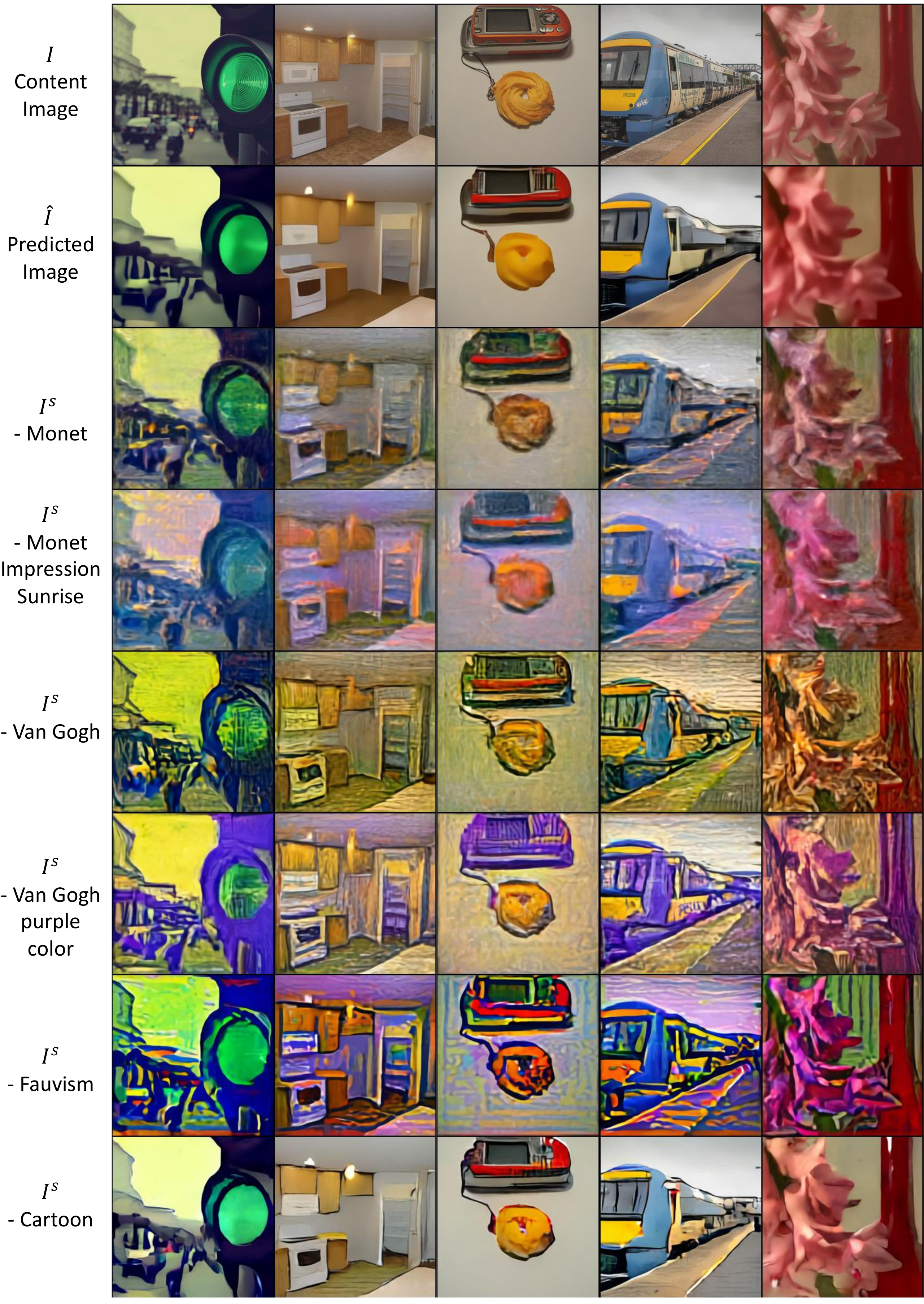}
    \end{center}
    \caption{Additional stylized results of $\mathtt{Styler}$DALLE-1.}
    \label{fig:apd3}
\end{figure*}

\begin{figure*}[ht]
    \begin{center}
    \includegraphics[width= .85\linewidth]{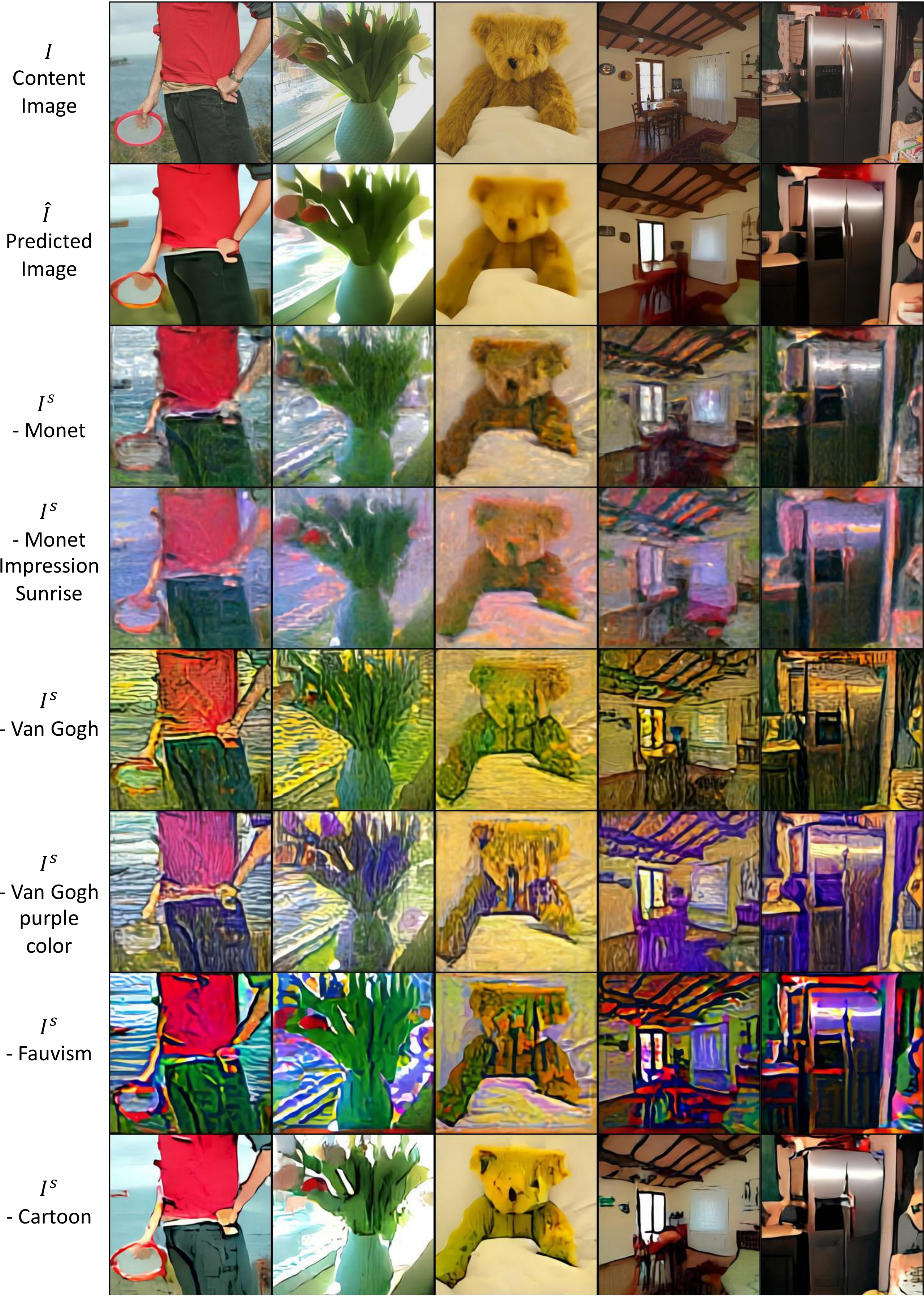}
    \end{center}
    \caption{Additional stylized results of $\mathtt{Styler}$DALLE-1.}
    \label{fig:apd4}
\end{figure*}

\paragraph{Additional Qualitative Results.}
We give more stylized results produced by $\mathtt{Styler}$DALLE-Ru in Fig.~\ref{fig:apdru3}, Fig.~\ref{fig:apdru2} and Fig.~\ref{fig:apdru1}, and $\mathtt{Styler}$DALLE-1 in Fig.~\ref{fig:apd2}, Fig.~\ref{fig:apd3} and Fig.~\ref{fig:apd4}, respectively.
In particular, we also show the intermediate results $\hat{I}$ (as in Fig.~\ref{fig.method}), which are generated with the output tokens using the model right after the self-supervised pre-training (and before the style-specific fine-tuning stage).
Similar to what we have concluded, both $\mathtt{Styler}$DALLE-1 and $\mathtt{Styler}$DALLE-Ru achieve distinctive and harmonious stylized results on various styles and images.
In addition, the differences between $\hat{I}$ and $I^s$ are significant. As shown, $\hat{I}$ is photo-realistic while $I^s$ presents varied brushstrokes, edges, and colors with respect to each style instruction, indicating that $\mathtt{Styler}$DALLE has been effectively fine-tuned with our language-guided rewards in the Reinforcement Learning stage.

By comparing the results of $\mathtt{Styler}$DALLE-1 and $\mathtt{Styler}$DALLE-Ru, although we draw the joint conclusions as above, we also see the differences between the two, resulting from the usage of different vector-quantized image tokenizers.
For example, $\mathtt{Styler}$DALLE-Ru achieves clearer stylized images, as it is implemented based on the VQGAN image tokenizer.
On the other hand, our method, i.e., $\mathtt{Styler}$DALLE has been proven effective on both  vector-quantized image tokenizers. 
It is reasonable to expect that the style transfer results can be further improved by using more advanced vector-quantized image tokenizers if they could be open-sourced.

\begin{figure*}[ht]
  \centering
  \subfigure[``3023"]{
  \begin{minipage}[]{\linewidth}
  \centering
    \includegraphics[width=.75\linewidth]{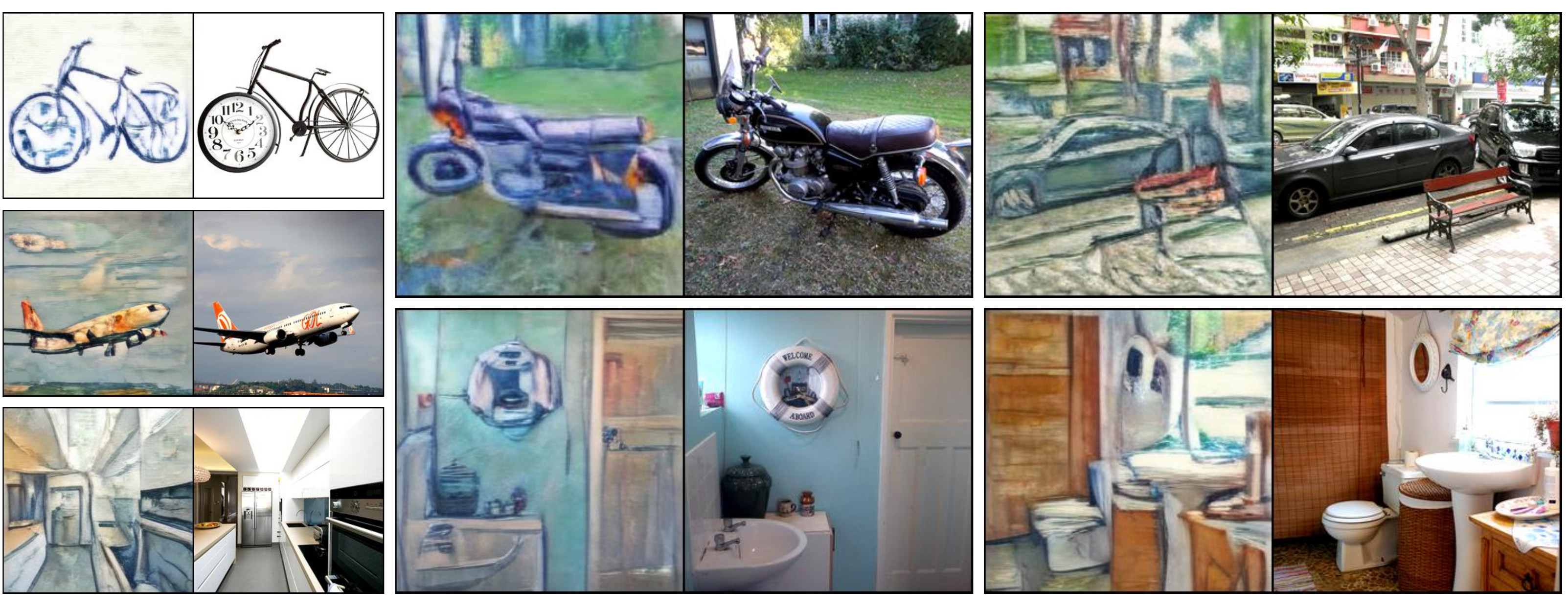}
    % \caption{``visionary art"}
\end{minipage}}
% \vspace{-0.2cm}
\subfigure[``a chill and sad Monet style painting"]{
\begin{minipage}[]{\linewidth}
\centering
    \includegraphics[width=.75\linewidth]{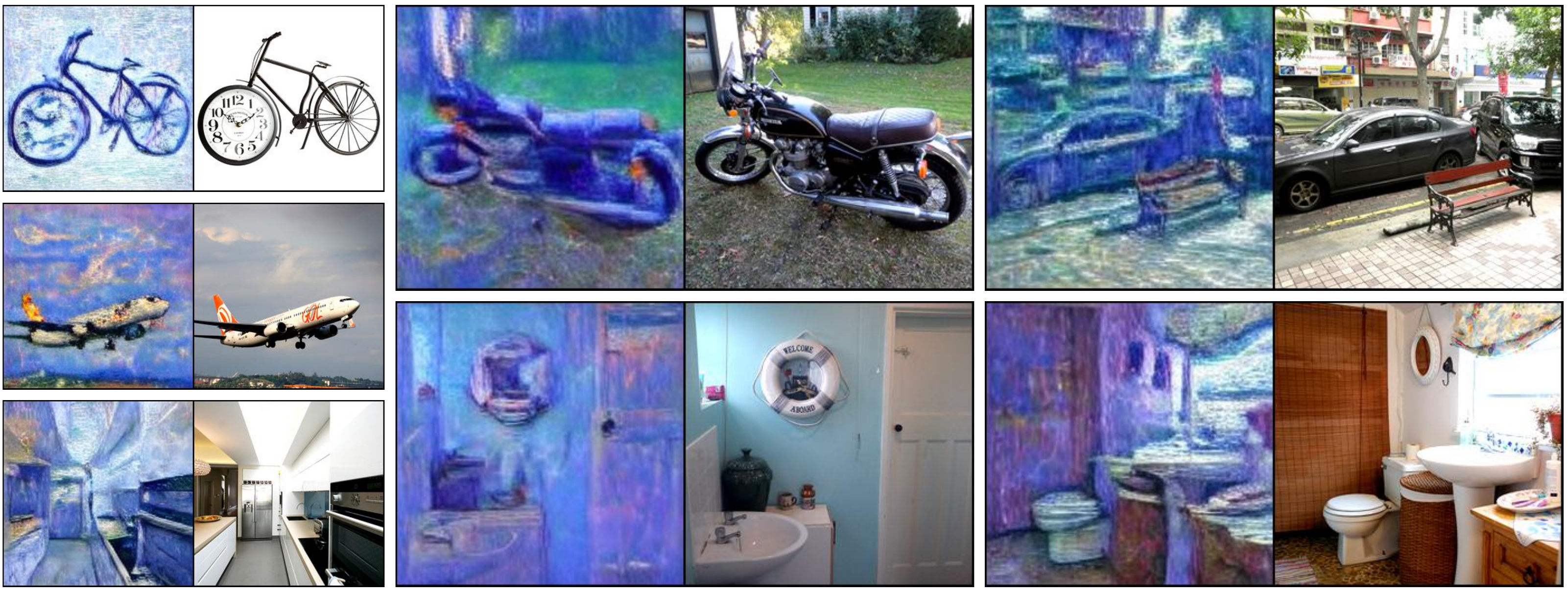}
    % \caption{``cartoon"}
  \end{minipage}}
\subfigure[``a rosy romantic relaxed Monet style painting"]{
\begin{minipage}[]{\linewidth}
\centering
    \includegraphics[width=.75\linewidth]{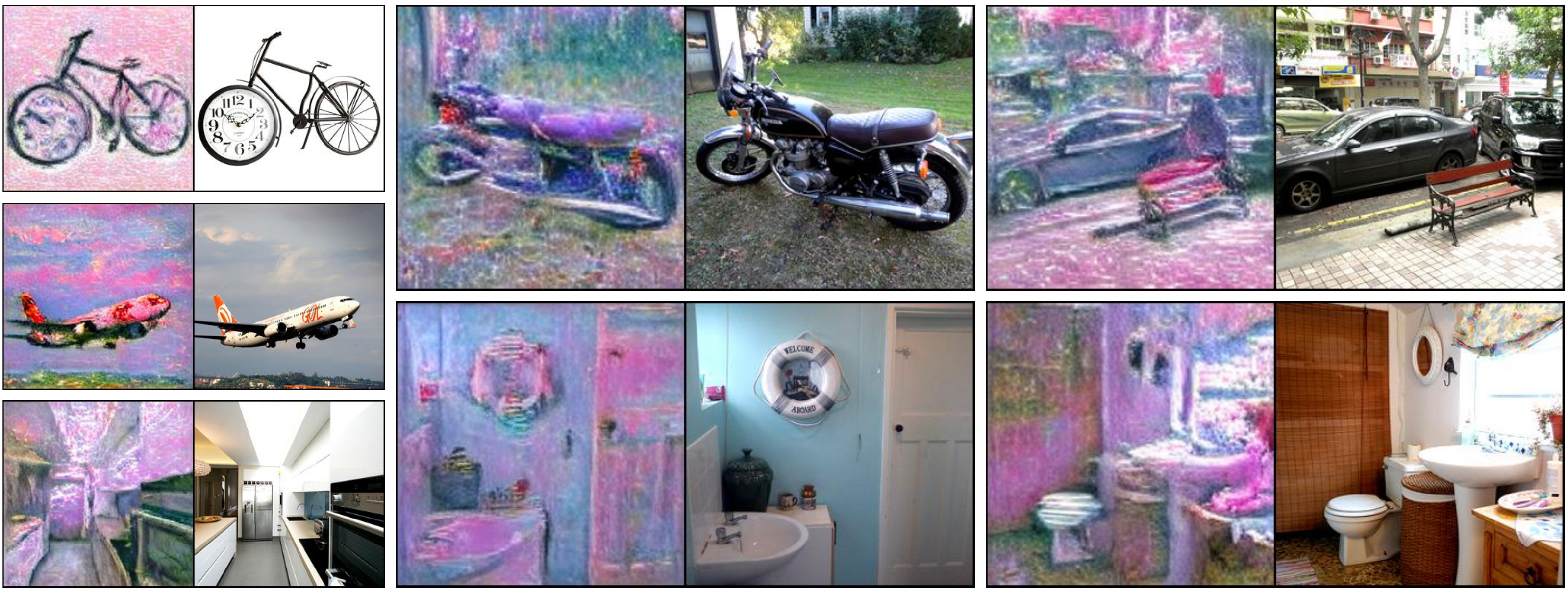}
    % \caption{``cartoon"}
  \end{minipage}}
\subfigure[``child drawing"]{
\begin{minipage}[]{\linewidth}
\centering
    \includegraphics[width=.75\linewidth]{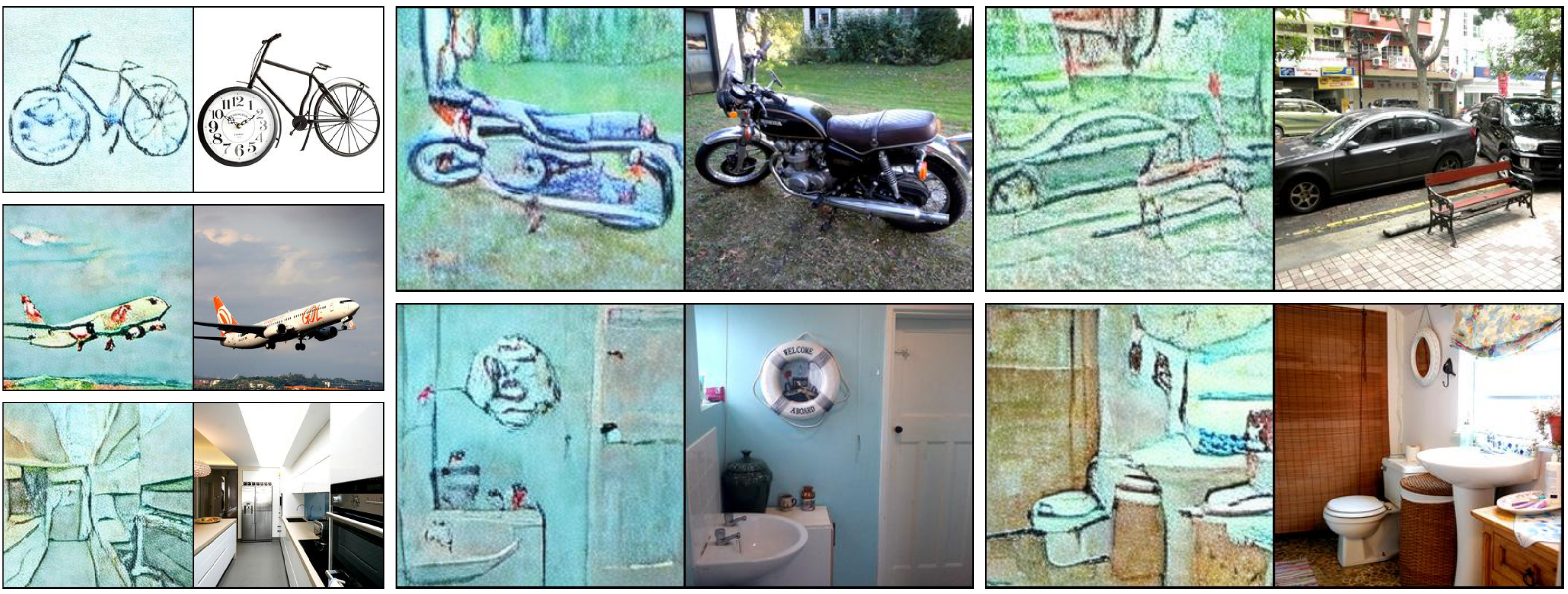}
    % \caption{``cartoon"}
  \end{minipage}}
  \caption{Non-cherry pick results on extra styles.}
  \label{fig:addd}
  % \vspace{-0.3cm}
\end{figure*}

In addition, we include non-cherry pick results on extra styles, i.e., ``3023", ``a  chill and sad Monet style painting", ``a rosy romantic relaxed Monet style painting" and ``child drawing", in Fig.~\ref{fig:addd}. These results come from $\mathtt{Styler}$DALLE-Ru.

\clearpage

\end{document}